\documentclass[lettersize,journal]{IEEEtran}
\usepackage{amsmath,amsfonts}
\usepackage{algorithmic}
\usepackage{algorithm}
\usepackage{array}
\usepackage[caption=false,font=normalsize,labelfont=sf,textfont=sf]{subfig}
\usepackage{textcomp}
\usepackage{stfloats}
\usepackage{url}
\usepackage{verbatim}
\usepackage{graphicx}
\usepackage{cite}
\usepackage{booktabs}
\usepackage{multirow}

\usepackage{xcolor}
\definecolor{myblack}{RGB}{192, 0, 0}
\definecolor{myblack}{RGB}{0, 0, 0}

\newcommand{\rev}[1]{\textcolor{myblack}{#1}}

\begin{document}

\title{Leveraging Color Naming for Image Enhancement}

\author{David Serrano-Lozano, Luis Herranz, Michael S. Brown, Javier Vazquez-Corral
\thanks{DSL and JVC are with the Universitat Autònoma de Barcelona, Spain and the Computer Vision Center, Barcelona, Spain. LH is with the Universidad Politécnica de Madrid, Spain. MSB is with the York University, Toronto, Canada.}}

\markboth{Journal of \LaTeX\ Class Files,~Vol.~14, No.~8, August~2021}%
{Serrano-Lozano \MakeLowercase{\textit{et al.}}: Leveraging Color Naming for Image Enhancement}


\maketitle

\begin{abstract}
Enhancing images to make them visually appealing is a persistent challenge in computer vision. Many deep-learning methods train models on paired datasets to replicate expert editing styles. However, these approaches struggle with two key issues: (1) interpretability and (2) a parametrization suitable for user adjustments. 
To address these challenges, we present NamedCurves+, an approach inspired by the concept of Color Naming, a universal set of familiar colors widely used in software tools for intuitive editing. Our method integrates color names into a learning-based framework, enabling global adjustments for each named color through tone curves. To address local image variations, we incorporate a transformer block that captures spatial dependencies, enabling context-aware edits across the image. NamedCurves+ enhances the retouching process's interpretability and supports user interaction, allowing flexible modifications of individual tone curves to refine the retouched image according to personal preferences. Extensive experiments on tasks such as image retouching, tone mapping, and exposure correction demonstrate that NamedCurves+ outperforms state-of-the-art methods. Notably, our approach is both explainable, as the tone curves explicitly represent how each color name contributes to the enhancement, and interactive, allowing users to customize the retouching process and achieve results tailored to their liking. Source code and models will be publicly available at: \url{https://namedcurves.github.io}.
\end{abstract}

\begin{IEEEkeywords}
Image Enhancement, Image Retouching, Color Editing, Color Naming
\end{IEEEkeywords}

\begin{figure*}[t]
    \centering

    \includegraphics[width=\linewidth]{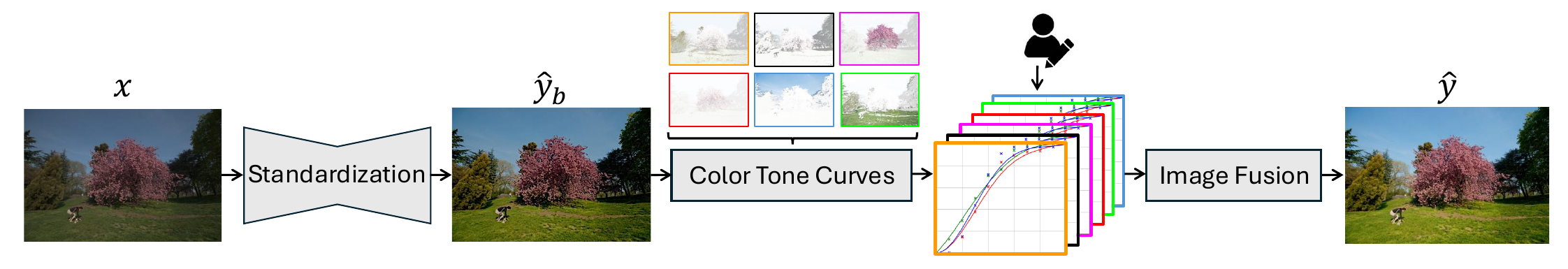}
    \makebox[.32\textwidth]{\footnotesize (a) Our framework}
    \vspace{2mm}
    


    \includegraphics[width=.22\linewidth]{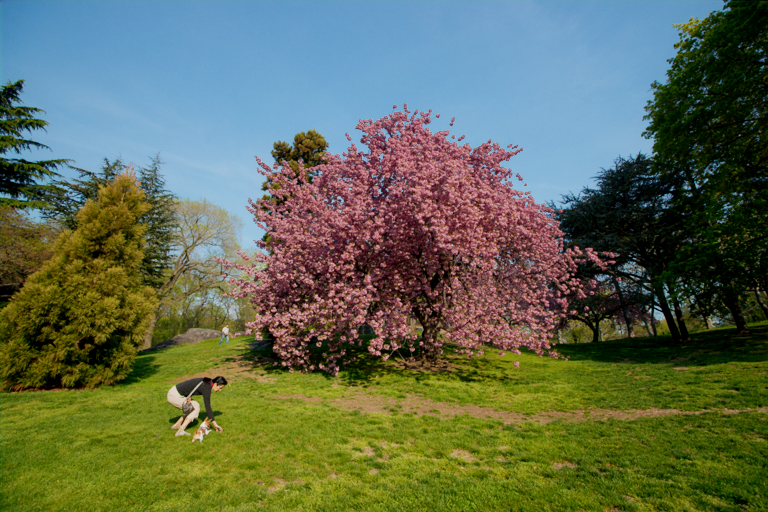}
    \includegraphics[width=.142\linewidth]{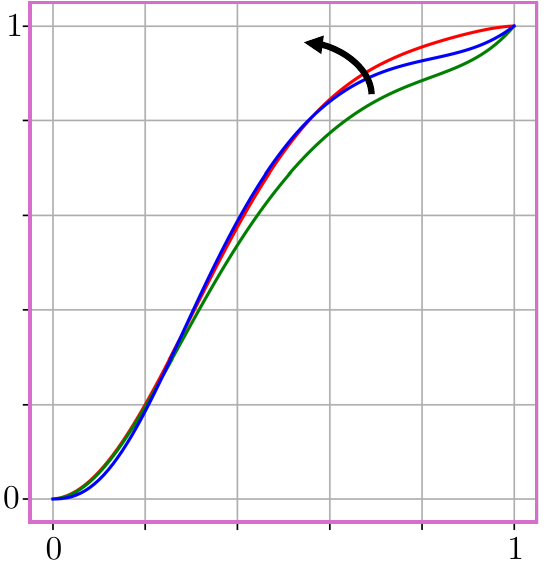}
    \includegraphics[width=.22\linewidth]{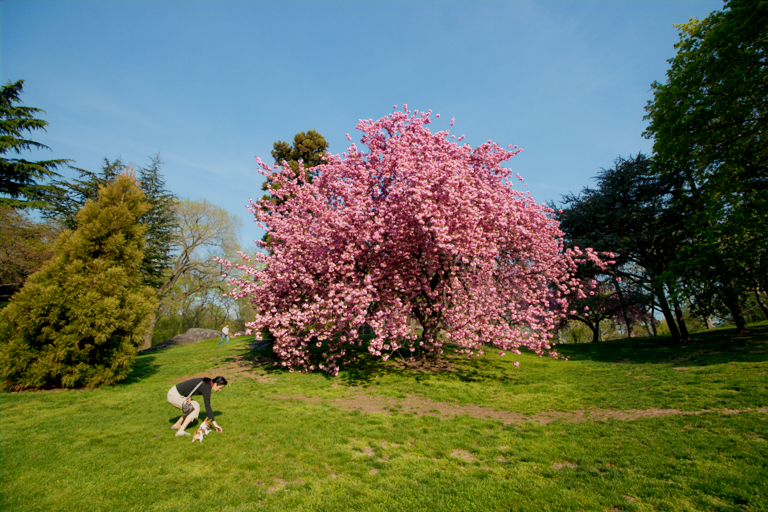}
    \includegraphics[width=.142\linewidth]{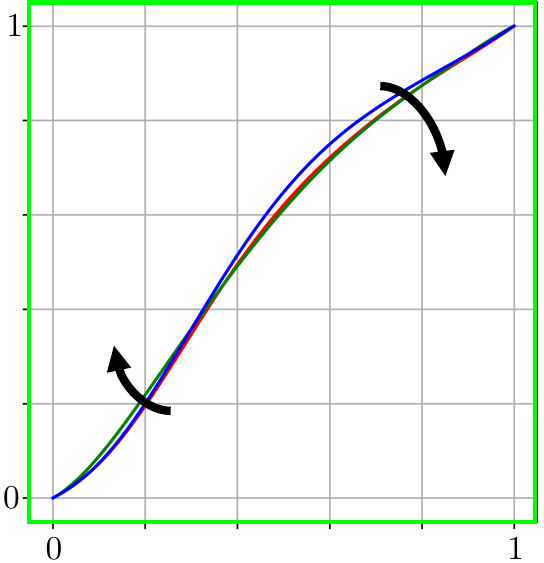}
    \includegraphics[width=.22\linewidth]{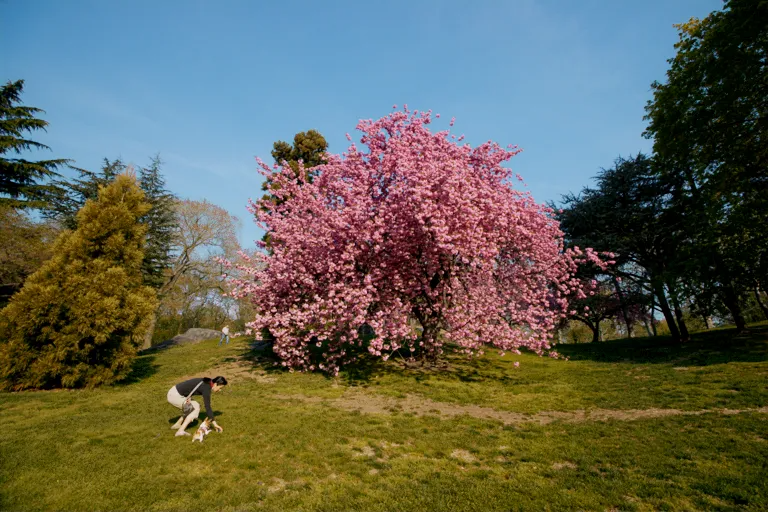}
    
    \makebox[.22\textwidth]{\footnotesize Output image $\hat{\text{y}}$}
    \makebox[.38\textwidth]{\footnotesize (b) Output image modifying the Pink-Purple curves}
    \makebox[.38\textwidth]{\footnotesize (c) Output image modifying the Green curves}
    \caption{Our framework (a) enhances images to achieve visually appealing results. Starting with the input image $x$, the image is first standardized by a UNet-like backbone $\hat{y_b}$. Next, the framework estimates a set of tone curves corresponding to each color name probability map. These tone curves are applied to generate intermediate images, which are then fused by a transformer block to produce the final enhanced image $\hat{y}$. NamedCurves+ supports user interaction, allowing flexible adjustments to global tone curves for each color name. In (b), modifying the highlights section of the Pink-Purple curve alters the pink tree, while in (c), reducing the shadows and highlights of the Green curves shifts green regions to a more brownish tone. The arrows indicate the direction of the curve adjustments. Note that the curve adjustments modify the initial enhanced version $\hat{y}$ without altering the model weights, which remain frozen.}
    \vspace{-4mm}
  \label{fig:teaser}
\end{figure*}

\vspace{-8pt}
\section{Introduction}
\IEEEPARstart{C}{olor} plays a vital role in photography, enhancing focal points, evoking emotions, and enriching storytelling. Whether through vibrant hues or subtle tones, understanding the importance of colors is crucial for photographers seeking to elicit specific responses. Despite significant advancements in camera technology, both amateur and professional photographers often resort to post-capture image enhancement to improve image quality. However, manual image editing poses challenges for those lacking expertise, time, or a well-developed aesthetic sense.

A potential solution to avoid manual adjustment lies in learning-based methods, where deep neural networks (DNN) model the editing style of skilled photographers or colorists. These models typically leverage paired datasets of raw and expert-retouched images. While effective in many scenarios, such approaches have critical limitations. Expert aesthetic sense may not always align with individual preferences or specific use cases, leaving certain users dissatisfied with the retouched results. Furthermore, existing methods often rely on approaches such as weighting multiple 3D Look-Up Tables~\cite{zeng2020learning, yang2022adaint, yang2022seplut, kim2025bilateral}, or employing end-to-end U-Net-like architectures~\cite{chen2018deep, ni2020towards, jiang2021enlightengan}, which offer limited interpretability and prevent user interaction. As a result, users cannot easily adjust the output to suit their specific needs. Compounding the issue, most image-retouching datasets are captured in controlled environments by skilled professionals, while real-world images are often taken by amateurs under varying lighting conditions. This creates significant variability in the amount of light reaching the camera sensor, further complicating the enhancement process. To bridge the gap between the intensive manual adjustments required by software tools and the limited controllability of current automated methods, we propose an approach that mimics the expert’s style while also allowing users to make subtle adjustments based on a predefined set of color curves.

It is interesting to consider the tools provided to the artists for performing the image editing. Many photo editing software applications (e.g., Adobe Photoshop~\cite{photoshop}) provide users with the ability to manipulate the image based on a small set of fixed colors (e.g., red, green, yellow, orange, blue, purple). Interestingly, the predefined colors selected by software tools are similar to those psychologists and linguists have found to be universal across languages~\cite{berlin1991basic}, a research topic often referred to as {\it color naming}.

This paper proposes leveraging pixel-wise color naming decomposition for image enhancement. Specifically, we introduce NamedCurves+, a learning-based framework comprising three main components. First, we employ a backbone model to standardize the input image into a canonical space. Next, using color naming, the image is decomposed into six color maps, each associated with a distinct set of Bezier-parametrized global tone curves. Finally, a transformer block combines the globally adjusted images to model local manipulations. Figure~\ref{fig:teaser} (a) illustrates the framework of our approach. By associating specific tone curves with each color name, NamedCurves+ offers both interpretability and interactivity. Interpretability is achieved thanks to the computation of different curves for each color name. Interactivity because users can directly manipulate the curves to alter a specific color, tailoring the output to their preferences. \rev{In detail, by modifying the tone curves of each color term, the user can obtain different outputs as the model enhances each color region differently. See Figure~\ref{fig:teaser} (b) and (c) to see different results when editing the pink-purple and green tone curves, respectively.} We evaluate NamedCurves+ on multiple tasks, including image retouching, tone mapping, and exposure correction, using benchmark datasets such as MIT-Adobe-5K~\cite{bychkovsky2011mit5k}, PPR10K~\cite{liang2021ppr10k}, MSEC~\cite{afifi2021msec}, and SICE~\cite{cai2018sice}. Our method achieves state-of-the-art performance both quantitatively and qualitatively, outperforming existing approaches.

\begin{figure}
    \centering
    \includegraphics[width=0.31\linewidth]{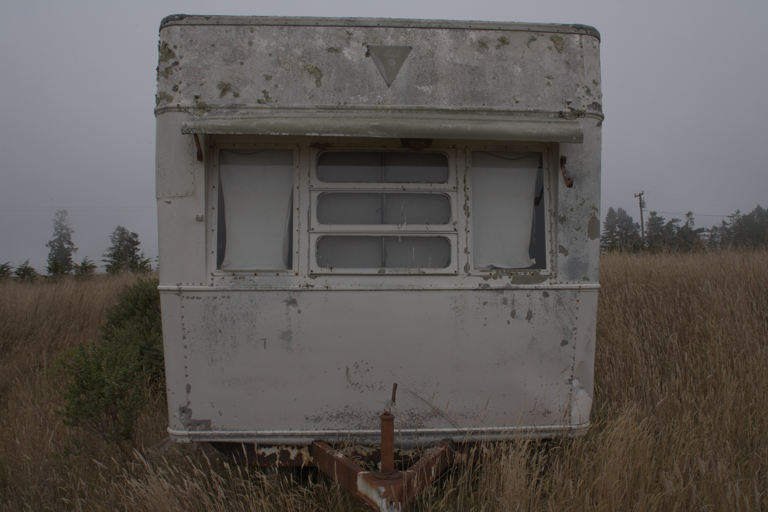}
    \includegraphics[width=0.31\linewidth]{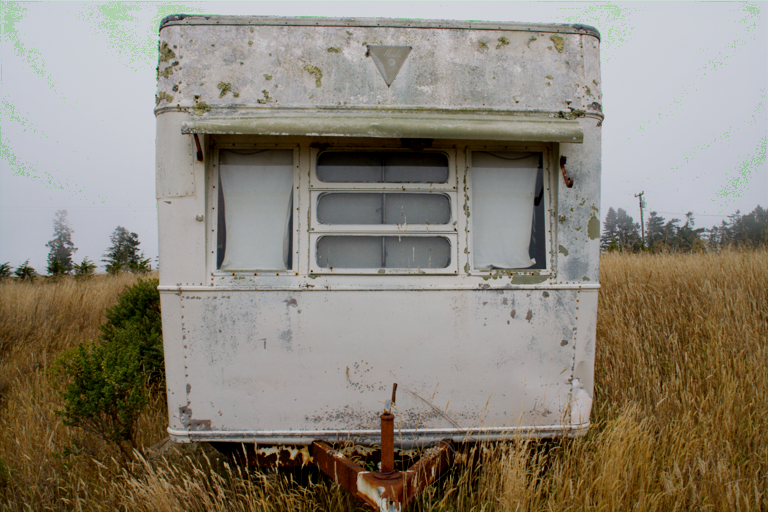}
    \includegraphics[width=0.31\linewidth]{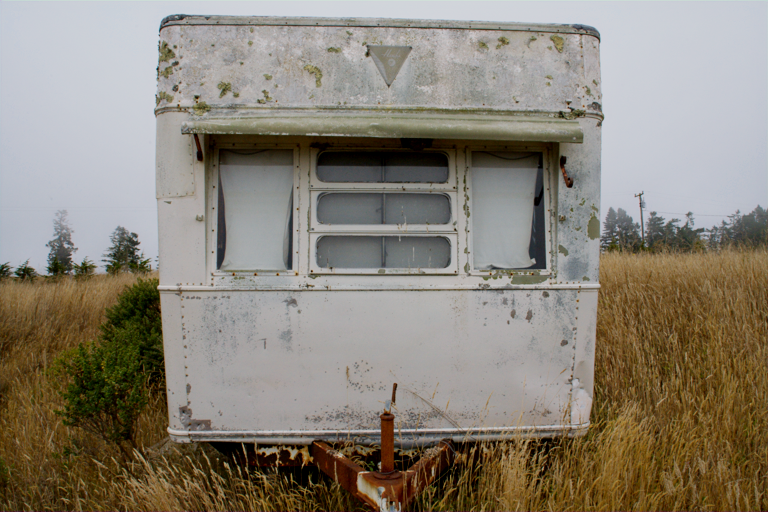}

    \vspace{1mm}

    \includegraphics[width=0.31\linewidth]{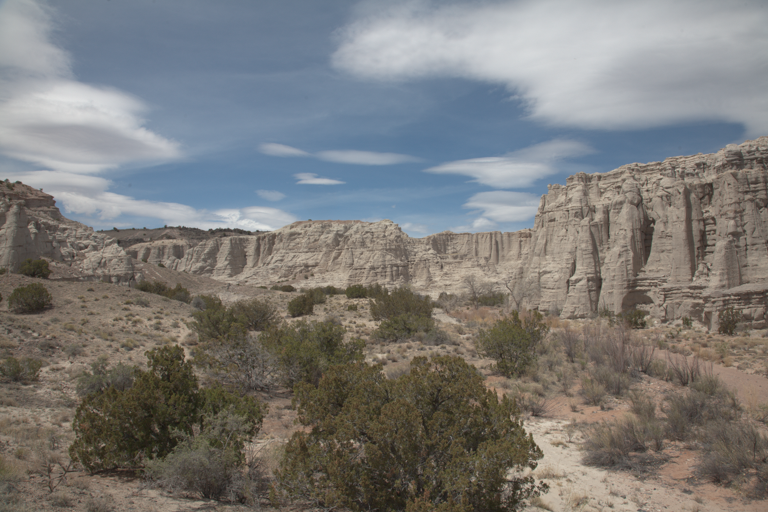}
    \includegraphics[width=0.31\linewidth]{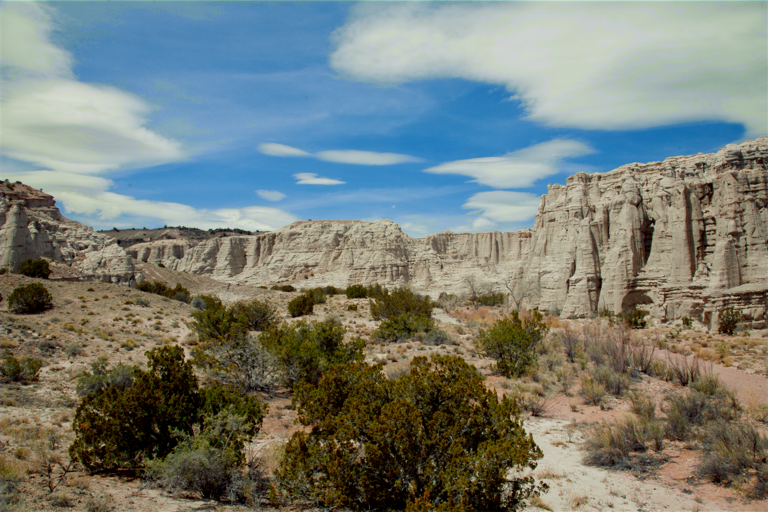}
    \includegraphics[width=0.31\linewidth]{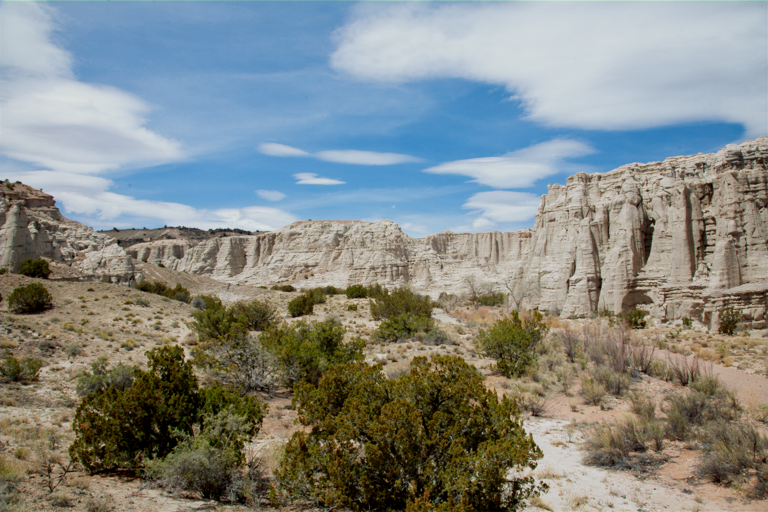}

    \makebox[.32\linewidth]{\footnotesize Input image}
    \makebox[.32\linewidth]{\footnotesize NamedCurves}
    \makebox[.32\linewidth]{\footnotesize NamedCurves+}
    \vspace{-2mm}
    \caption{\rev{Examples where NamedCurves struggled with. It generates halo artifacts in color boundaries and brittleness when color naming probabilities change drastically.}}
    \label{fig:motivation}
\end{figure}

This work extends our previous conference paper~\cite{serrano2025namedcurves}, where we introduced a learning-based method that leverages color naming and tone curves for end-to-end image retouching. In this extended version, we present NamedCurves+ as a more general image enhancement framework with improved performance and efficiency across multiple tasks, including tone mapping and exposure correction. We also strengthen interpretability and interactivity by decomposing the image into a set of predefined color names and a color-conditioned tone curves that users can directly edit to steer the final result without retraining.
\rev{In particular, we make the following contributions: (i) we redesign the image fusion component using an efficient transformer block, improving quality and reducing artifacts while maintaining fast inference; (ii) we extend the framework beyond retouching to tone mapping and exposure correction, achieving state-of-the-art results on standard benchmarks; and (iii) we demonstrate user control by modifying the tone curves associated with each color name to obtain customized enhancements.
Finally, we address a key limitation of our previous architecture: its fusion mechanism could be semantically brittle when color-term probabilities overlap or change smoothly, leading to halo artifacts at color boundaries. Figure~\ref{fig:motivation} highlights two representative failure cases of NamedCurves: in the first example, haloing appears near the top corners when pixels have high probability for multiple color terms, and in the second example, inconsistencies are visible in the cloud region. By contrast, NamedCurves+ mitigates these artifacts through the proposed end-to-end fusion design, producing cleaner and more visually consistent enhanced images.}

\section{Related Work}
\textbf{Color Naming.} Color naming is crucial for product design, photography, and vision research \cite{xue2023integrating, xue2024palette, regier2007color, szafir2017modeling, bahng2018coloring}. Berlin and Kay \cite{berlin1991basic} conducted a study on the basic color lexicon across various languages and discovered universal semantics. Their seminal analysis showed that the evolution of basic color vocabularies is influenced by visual physiology, which limits the possible composite categories to a small number of those. The 11 color names found that most societies and cultures share are: \textit{orange}, \textit{brown}, \textit{yellow}, \textit{white}, \textit{grey}, \textit{black}, \textit{pink}, \textit{purple}, \textit{red}, \textit{green} and \textit{blue}. 

\begin{figure}[t]
    \centering
    \includegraphics[width=\linewidth]{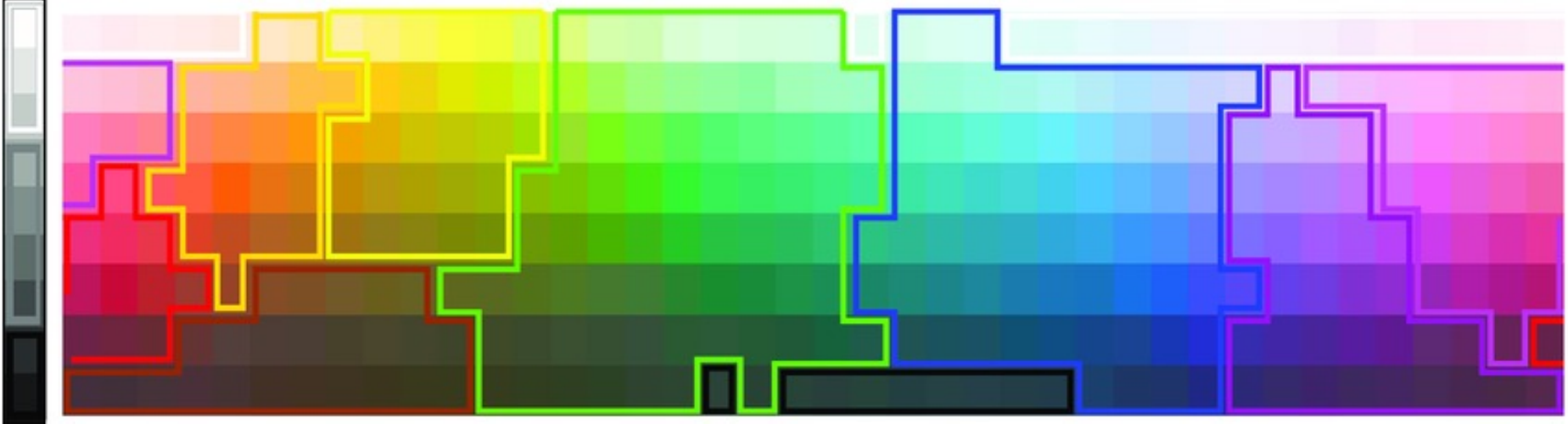}
    \vspace{-6mm}
    \caption{Van de Weijer et al. \cite{van2009learning} color names grouped in the Munsell color array. The color names are \textit{white}, \textit{grey}, \textit{black}, \textit{red}, \textit{brown}, \textit{orange}, \textit{yellow}, \textit{green}, \textit{blue}, \textit{purple} and \textit{pink}.}
    \vspace{-4mm}
    \label{fig:munsell-color-array}
\end{figure}

Following Berlin and Kay's research, different studies (e.g., \cite{van2009learning, benavente2008parametric, yu2018weakly, parraga2016nice})  aimed at predicting the boundaries between each color name. For example, Figure~\ref{fig:munsell-color-array} shows the standard Munsell color array using Van de Weijer et al.~\cite{van2009learning} color classification based on color naming. 

\begin{figure*}[t]
    \centering
    \includegraphics[width=\linewidth]{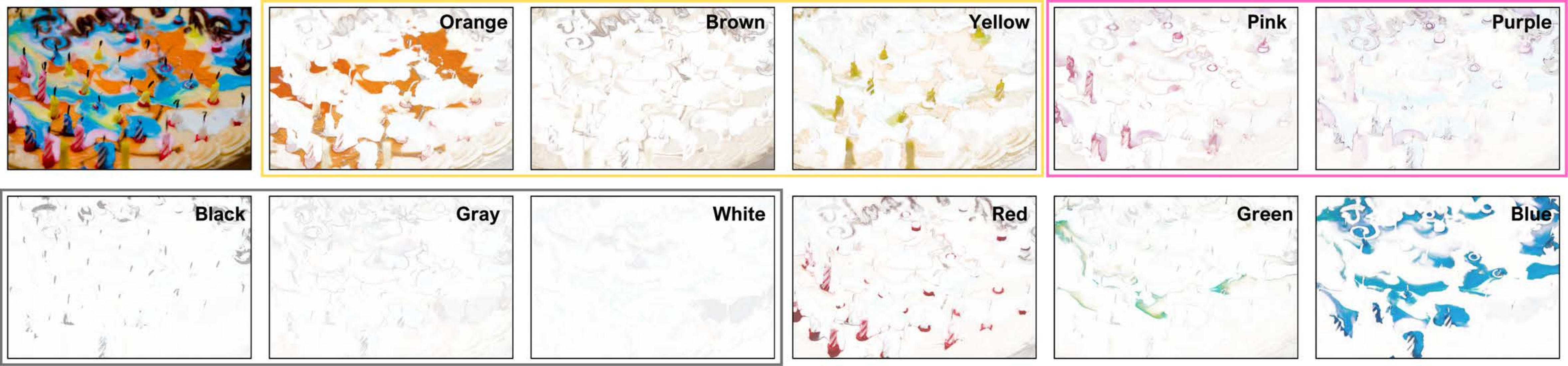}
    \caption{Van de Weijer et al. \cite{van2009learning} color naming method applied pixel-wise to the top-left image. The other 11 images show the 11 probability color names maps. Each color is displayed with a different map to aid visualization. Note that some linguistic color names share approximately the same hue and only differ in intensity--- e.g., \textit{pink} and \textit{purple}. As tone curves are defined for all the intensity range we group: \textit{orange-brown-yellow}, \textit{pink-purple}, and \textit{white-grey-black}. This grouping is represented by the boxes.}
    \label{fig:color-naming-image}
\end{figure*}

These methods work as follows. Given an RGB value in the sRGB color space, color naming methods produce an 11-d vector corresponding to the probability of the RGB value belonging to the specific color names listed above. This is visualized in Figure \ref{fig:color-naming-image}, where we show an original image and the 11-probability maps coded with a color map to aid visualization.

\textbf{Learned Image Enhancement.} The need to provide users with tools to allow easy image enhancement has grown significantly due to the ease of smartphone photo-taking.  Initially, histogram equalization was a primary method for enhancing contrast in images \cite{stark2000adaptive, reza2004realization}. Subsequently, local operators \cite{aubry2014fast, durand2002fast} and color correction techniques based on color constancy \cite{van2007edge} were introduced. Since the introduction of the MIT-Adobe-5K dataset by Bychkovsky et al. \cite{bychkovsky2011mit5k}, which contains 5,000 images retouched by five experts, data-driven methods have emerged as one of the preferred means to improve image quality. 

Lookup tables (LUTs) represent a widely used method for image manipulation, typically manually tuned and fixed in camera imaging pipelines or photo editing tools. Zeng et al.~\cite{zeng2020learning} proposed 3DLUT, a method to learn these 3D LUTs from annotated data with a small convolutional neural network. Building upon this, Yang et al.~\cite{yang2022adaint} proposed a mechanism to achieve a more flexible sampling by learning the non-uniform sampling intervals, Yang et al.~\cite{yang2022seplut} proposed mixing 1DLUTs and 3DLUTs to obtain a good trade off between memory footprint and transformation capabilities, and Kim et al.~\cite{kim2025bilateral} proposed including bilateral grids to incorporate spatial information. Wang et al. \cite{wang2021real} also presented a modification of 3DLUT that incorporates spatial data extracted with a shallow UNet architecture.

Conversely, image-to-image methods estimate directly a mapping to modify the input images without intermediate steps. Generative adversarial networks (GANs) are frequently employed for such tasks. Chen et al. \cite{chen2018deep}, Ni et al. \cite{ni2020towards}, and Jiang et al. \cite{jiang2021enlightengan} proposed unpaired learning schemes using single GANs to estimate enhanced versions of input images directly.

However, the group of methods closer to our approach are the ones that involve estimating intermediate or physical parameters. \rev{DeepLPF~\cite{moran2020deeplpf} and RsfNet~\cite{ouyang2023rsfnet}, inspired by retouching software tools, estimate elliptical, gradual, and polynomial filters for local image editing and region-specific filters, respectively. They showed promising results, but editing only locally the images introduced artifacts into the final image.} Guo et al. \cite{guo2020zero} proposed Zero-DCE, the first method to formulate low-light image enhancement as a curve estimation problem. Their deep network estimates pixel-wise curves to modify the dynamic range of input images. This groundbreaking work influenced subsequent methods, such as CURL \cite{moran2021curl}, which estimates piecewise linear curves for HSV, RGB, and CIELab color spaces; FlexiCurve \cite{li2023flexicurve}, which estimates sets of piecewise curves and blends them via a Transformer, Bianco et al.~\cite{bianco2020personalized} which estimates the global curves to retouch an image, and LTMNet \cite{zhao2022learning} that learns a grid of tone curves to both locally and globally enhance an image. Many previous methods~\cite{yang2022adaint, yang2022seplut, kim2025bilateral, zeng2020learning} have also been evaluated on learned tone mapping using the widely used MIT-Adobe-5K dataset~\cite{bychkovsky2011mit5k}. In contrast to image retouching, the images are in 16-bit CIE XYZ format and the outputs are sRGB images.

As in previous methods~\cite{moran2021curl, li2023flexicurve, bianco2020personalized, zhao2022learning}, we use tone curves to manipulate images. Our main novelty is that we leverage color naming decomposition to disentangle distinct colors and later employ a transformer-based fusion scheme to emulate the image editing style of an expert. While previous methods may appear to offer interpretability through tone curves, this is often not the case. Manipulating images in different spaces or concatenating various tone curve sets undermines interpretability. In contrast, our approach disentangles different color names, enabling a clear interpretation of how the model manipulates each color in the image.

\textbf{Learned Exposure Correction.} Incorrect exposure remains a significant source of errors in camera-based imaging, arising from excessively long exposure times (overexposure) or excessively short ones (underexposure). Existing methods have proposed various specialized approaches to address exposure correction. For instance, Afifi et al.\cite{afifi2021msec} formulate the problem as a combination of color and detail enhancement using a two-step DNN. Nsampi et al.\cite{nsampi2021cmec} introduce a deep feature matching loss to enable exposure-invariant feature representation, while Huang et al.~\cite{huang2022fecnet} propose lightness reconstruction using two Fourier-based DNNs: one for amplitude and another for phase.

This paper extends from our conference paper~\cite{serrano2025namedcurves}, where we introduced the use of color naming and tone curves for end-to-end image retouching. NamedCurves+ addresses the broader challenge of developing a generic image enhancement framework. Specifically, we improve the architecture by incorporating a Transformer block to blend the globally adjusted images while modeling the spatial dependencies between color-specific adjustments. We evaluate our framework across three tasks: image retouching, tone mapping, and exposure correction, using four well-established datasets, demonstrating consistent improvements over state-of-the-art methods and our prior approach. Additionally, due to its parameterized design, NamedCurves+ allows users to fine-tune individual color tone curves for personalized control over the enhanced image. Finally, we thoroughly analyze the advantages of color naming, highlighting its role in improving both interpretability and user interaction in image enhancement.

\section{Method}\label{sec:method}
Figure \ref{fig:teaser} (a) shows an overview of our proposed method, NamedCurves+. Our method aims to enhance a low-quality RGB input image $x$, by a learned model that outputs an enhanced version $\hat{y}$. This resulting enhanced image is expected to be close to the expert-retouched image $y$, based on some objective function $L$.  

Our method consists of four main components, which are detailed in the following sections, including the loss function used for optimizing the framework. The approach first applies a DNN backbone that standardizes the input image into a canonical latent space. Next, we use color naming to decompose the image into six color maps. After color naming decomposition, a neural network learns a set of Bezier tone curves to manipulate each color map globally. Finally, a transformer module combines the edited images to achieve local editing effects.


    




\subsection{Backbone}\label{subsec:backbone}
One challenge faced by learning-based image enhancement methods is that input images, $x$, can be captured using different cameras with different settings and under different lighting conditions. This may impact our ability for consistent color naming.  Similar to the method by Moran et al.~\cite{moran2020deeplpf}, we use a UNet-like backbone to standardize the input images. 

Our backbone is inspired by LPIENet \cite{conde2023perceptual} architecture. We use MobileNet~\cite{howard2017mobilenets} layers (\texttt{Conv-DWConv-eLU}) and a CBAM module \cite{woo2018cbam}---a combination of spatial and channel attention. The backbone consists of three encoder blocks and two decoder blocks connected by multi-resolution skip connections. Each encoder block consists of the following: two MobileNet layers, a CBAM attention block, and a max-pooling layer. The decoder blocks follow the same structure except for the max-pooling layers that are replaced by bilinear upsampling layers. The multi-resolution skip connections consist of three parallel branches of convolutional layers with different dilation rates. Two of the paths consist of two \texttt{Conv-LeakyReLU} blocks to extract local information, while the other path consists of three \texttt{Conv-LeakyReLU-MaxPooling} blocks, an \texttt{AveragePooling} and a \texttt{LinearLayer} to extract global information. As in \cite{moran2021curl, marnerides2018expandnet, gharbi2017deep}, we found that skip connections at different resolutions improve the performance against backbones with simple skip connections. 

\subsection{Color Naming}\label{subsec:method_naming}

We aim to decompose the standardized image $\hat{y}_b$ into a set of likelihood color maps to focus different branches of the model. Motivated by the importance of memory colors --- the green of the grass or the blue of the sky --- in aesthetics \cite{shing2010influences,topfer2000quantitative} we decided to use Color Naming, a perceptually-based color decomposition.

We used the color naming model from Van de Weijer et al. \cite{van2009learning} to obtain the probability maps for each color name. This model inputs an sRGB color value and outputs the probability of this color belonging to each $11$ color naming categories, namely \textit{red, blue, green, yellow, pink, purple, orange, brown, white, grey, black}. When applied to an image, the model operates for each pixel, which returns a set of probability maps.

We note that some linguistic color names share similar hues but only differ in intensity. For example, orange and brown or pink and purple. As we manipulate the different colors for all intensity ranges, it is beneficial to group these colors together. To this end, we reduce the set of $11$ probability maps to just $6$ by grouping \textit{orange-brown-yellow}, \textit{pink-purple}, and \textit{white-grey-black} (referring to this last one as \textit{achromatic}). The combined map for these cases is just the addition of the individual maps, and therefore, they are still probabilities (the sum of all the maps for a specific pixel is $1$).

\begin{figure}[t]
    \centering
    \includegraphics[width=.32\linewidth]{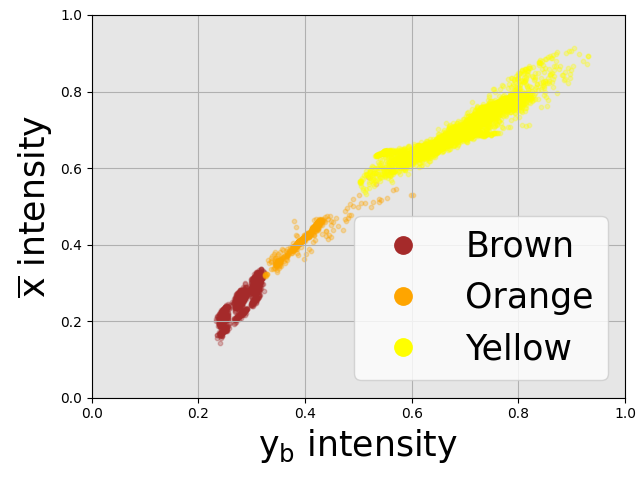}
    \includegraphics[width=.32\linewidth]{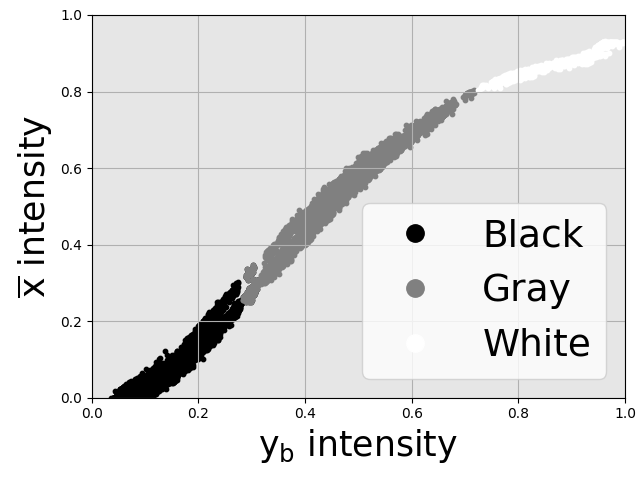}
    \includegraphics[width=.32\linewidth]{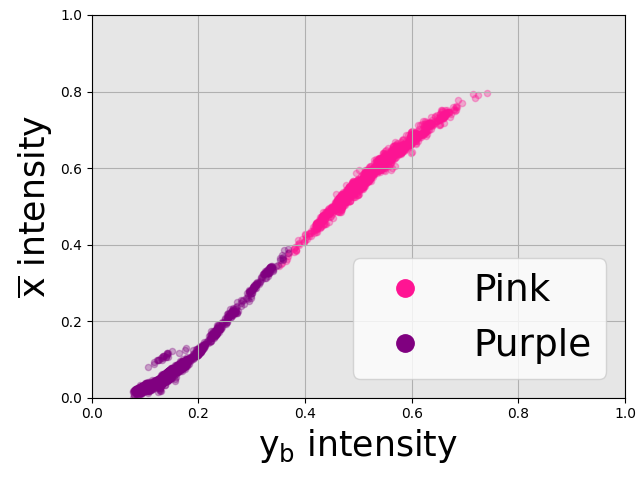}
    \makebox[.015\linewidth]{}
    \makebox[.315\linewidth]{\footnotesize (a) Orange-Brown-Yellow}
    \makebox[.315\linewidth]{\footnotesize (b) Achromatic}
    \makebox[.315\linewidth]{\footnotesize (c) Pink-Purple}
    \caption{Grouped color names with respect to the intensity value. Pixels with color name probability greater than 0.5 are plotted.}
  \label{fig:plot_group}
\end{figure}

In Figure~\ref{fig:plot_group}, we visually show the aforementioned reason for reducing the number of color naming channels to just 6, i.e. these colors have similar hues at different intensities. We illustrate plots depicting the relationship between input and output intensity values for pixels with the probability of belonging to each specific color greater than 0.5. For example, in the case of orange-brown-yellow, brown is only present at low intensities, while orange dominates at mid-intensities and yellow at top intensities. The same analysis extends to the other joined color channels. Our method aims to learn a curve to be applied at all the intensity levels. Thus, incorporating information spanning all the intensity levels is beneficial. In Figure \ref{fig:color-naming-image}, the color names grouped are shown in boxes. 

\begin{figure}[t!]
    \centering
    \includegraphics[width=\linewidth]{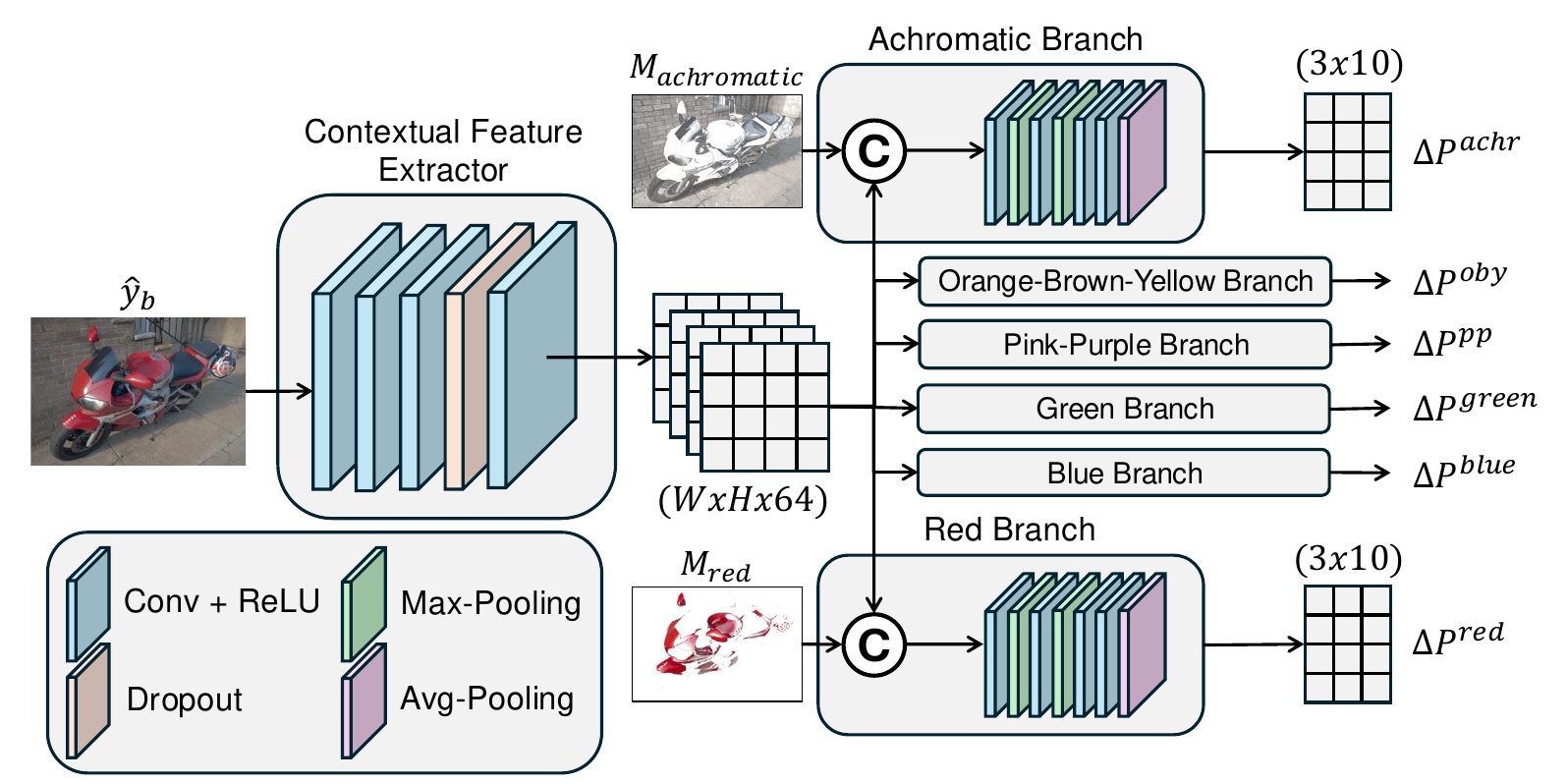}
    \caption{Estimation of the Bezier tone curves. First, we extract 64-D convolutional feature maps from $\hat{y}_b$ using mainly 4 \texttt{Conv-ReLU} blocks. Then, each specific color name map is concatenated to the feature maps and passed through 4 \texttt{Conv-ReLU-MaxPooling} and a final \texttt{AveragePooling} layer. The module's output is $\Delta P$, and the unnormalized control points increments.}
    \label{fig:bcpe}
\end{figure}

\subsection{Bezier Tone Curves Estimation}
Similar to prior works \cite{guo2020zero, moran2021curl, li2023flexicurve}, we leverage tone curves to remap the shadows, midtones, and highlights of the image $\hat{y}_b$ conditioned by the color naming probability maps. Tone curves represent global intensity adjustments from the input level to the output level. The curves are applied pixel-wise in each color channel as 1D Look-Up Tables. Bennet and Finlayson~\cite{bennett2023simplifying} demonstrated that tone adjustments are typically simple curves for a large dataset of enhanced images or can be well-approximated as such. We use Bezier curve parametrization to generate smooth and continuous tone curves from discrete control points.

Specifically, we aim to estimate one global curve for each RGB channel $c$ and color name $n$. Each curve is parameterized by $M$ control points. We evenly distribute these control points along the input axis, with the first control point fixed at $(0, 0)$. Consequently, we only need to estimate the control points' output axis values instead of the two-point coordinates, resulting in $M\!-\!1$ parameters per curve. Thus, the Bezier formulation $B^{n, c}$ of a curve can be expressed as:
\begin{equation}
    B^{n, c}(i) = \sum_{m=0}^{M\!-\!1} P_m^{n, c} \binom{M\!-\!1}{m} (1-i)^{(M-1-m)} i^{m},
    \label{eq:Bezier}
\end{equation}
where $i \in [0, 1] $ is an image channel pixel, $P_m^{n, c}$ denotes the $m$-th control point for the color name $n$ and channel $c$, and $M$ is the total number of control points.

Figure \ref{fig:bcpe} illustrates the estimation of the control points that define the Bezier tone curves of an image, comprising two distinct blocks: the contextual feature extractor and $6$ color naming branches. The contextual feature extractor primarily consists of $4$ \texttt{Conv-ReLU} blocks and a \texttt{Dropout}, while the color naming branches consist of $4$ \texttt{Conv-ReLU-MaxPooling} blocks and a final \texttt{AveragePooling} layer to manage the variable sizes of the images.

The contextual feature extractor computes a $64$-d convolutional feature map from the standardized image $\hat{y}_b$. Each color naming branch receives as input a concatenation of the $64$-d feature maps and the corresponding color naming probability map. The output of the branch for color name $n$ consists of three sets of $M$ increments $\left(\Delta P^{n,c}_m\right)_{m=1}^{M}$, each set corresponding to a curve for a given color channel $c$. These increments $\Delta P^{n,c}_m$ do not directly represent the control points as we impose two different constraints. First, to make the curves monotonically increasing functions, we define $\Delta P^{n,c}_m$ as positive increments relative to the previous point. Second, we normalize $\Delta P^{n,c}_m$, ensuring the total increment between the first and last points is 1 and, thus, placing the last point at $(1, 1)$. Consequently, we compute the control points $P^{n,c}_m$ as the accumulated sum of the normalized $\Delta P^{n,c}_m$. This can be formulated as:

\begin{equation}
    P_m^{n, c} = \frac{1}{S^{n,c}} \sum_{k=1}^{m} \Delta P_k^{n, c},
    \label{eq:controlpoints}
\end{equation}
where $S^{n,c}= \sum_{k=1}^{M} \Delta P_k^{n, c}$.

\begin{figure}[t!]
    \centering
    \includegraphics[width=\linewidth]{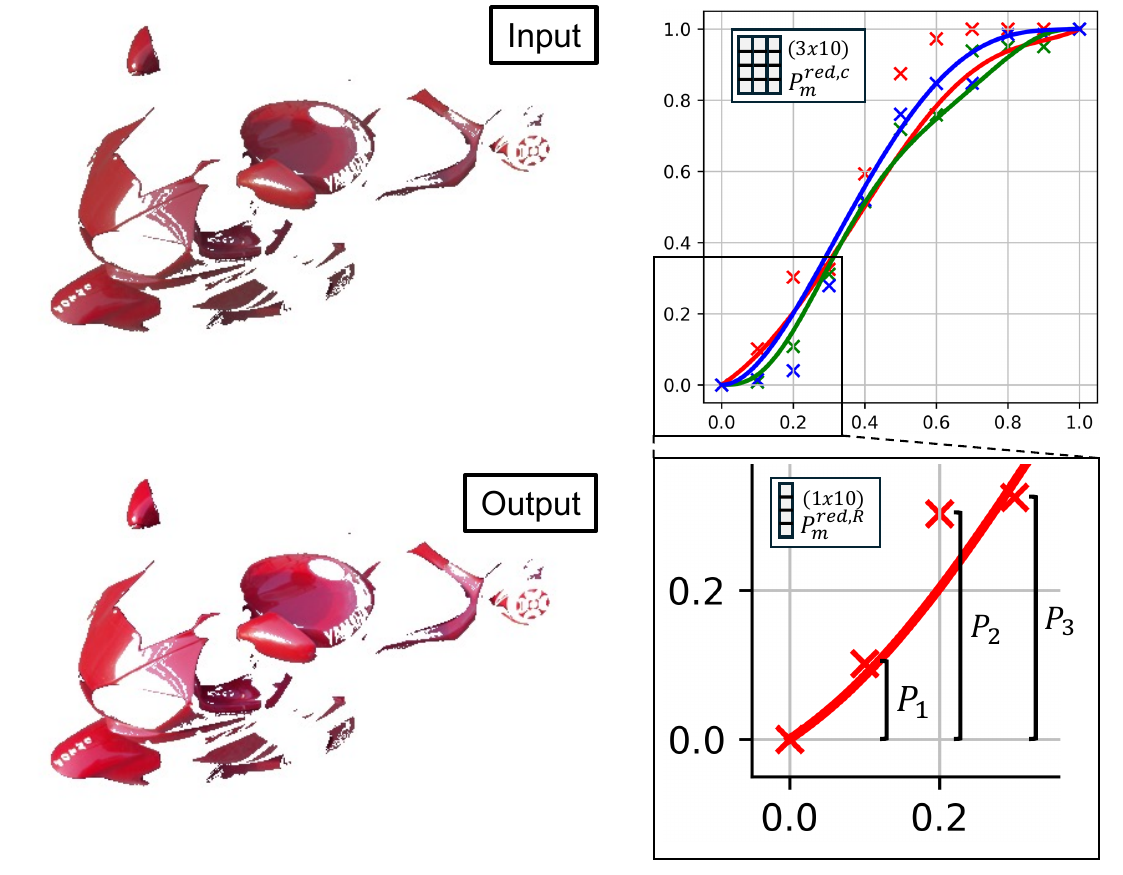}
    \caption{An example of a Bezier curve (Red branch). The first column shows the input and edited image pixels in which the red color name branch focuses. The center plot shows the tone RGB curves learned for the red color name, while the right plot displays a zoomed-in view of the first four control points of the red channel.}
    \vspace{-4mm}
    \label{fig:controlpoints-plot}
\end{figure}

Figure \ref{fig:controlpoints-plot} illustrates an example of a Bezier curve. The left column shows the input and output image pixels with a higher red color name probability than 0.2. The top-right plot shows the tone curves learned for the red color name, while the bottom-right plot displays a zoomed-in view of the first four control points of the R channel. Note how the control points are fixed and evenly distributed along the input axis, while $P^{n,c}_{m}$ define the output axis value and, thus, the curvature of the tonal curve. The six sets of Bezier curves learned are applied pixel-wise to the entire standardized image $\hat{y}_b$, yielding six distinct globally adjusted images.

\subsection{Transformer-based Image Fusion}
Tone curves enable interpretable global color manipulation, but they are not sufficient to reproduce the localized, content-dependent adjustments typically performed by experts (e.g., selectively enhancing a subject while keeping the background unchanged). To capture these local editing effects, we introduce a transformer-based fusion block that combines the six globally adjusted images into a single enhanced output.

\begin{figure*}[t!]
    \centering
    \includegraphics[width=\linewidth]{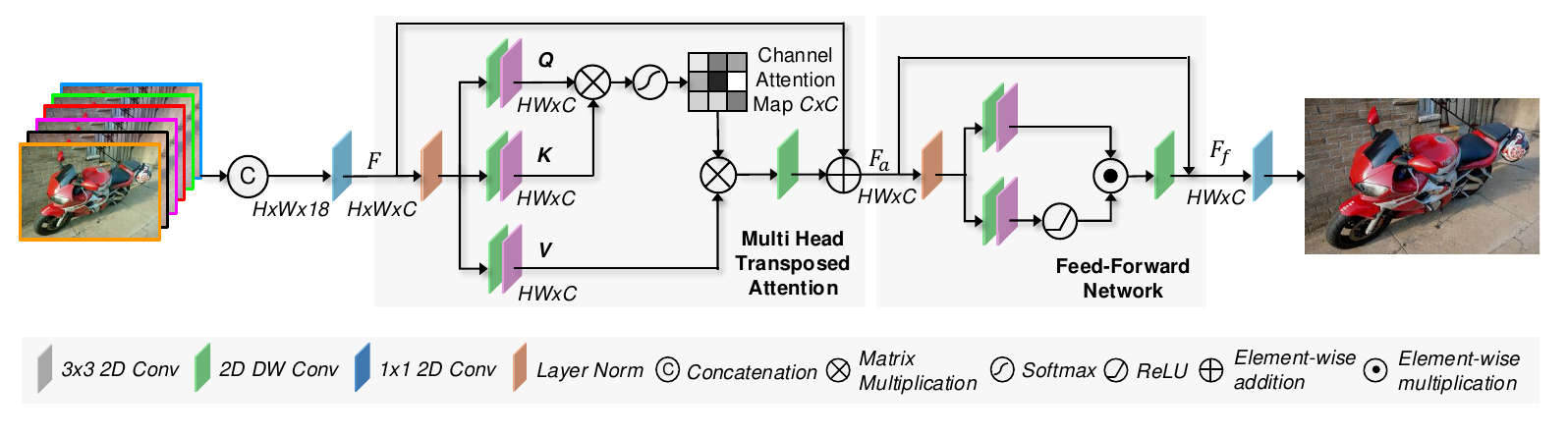}
    \vspace{-10mm}
    \caption{Overview of the transformer block. The globally-adjusted images, each one conditioned by a color term, are concatenated and passed through a point-wise convolution layer. The resulting features, $F$, are then processed by a Multi-Head Transposed Attention module and refined by a Feed-Forward Network, culminating in a three-channel output that forms the final enhanced image.}
    \label{fig:transformer}
\end{figure*}

\rev{
\textbf{Limitations of the previous fusion mechanism.}
In our previous approach, NamedCurves, we processed the six globally adjusted images independently with a spatial self-attention module and then averaged the results using the color naming probabilities as weights. Concretely, each globally adjusted image provided the {\it key} and {\it value}, while the standardized image acted as the {\it query}. We first reduced the spatial resolution with two \texttt{Conv-ReLU-MaxPooling} blocks, producing feature maps with a broad receptive field; in particular, $Q, K, V \in \mathbb{R}^{W/8 \times H/8 \times 16}$. The output of each attention module, $\hat{y}_i$, can be written as:
\begin{equation}
    \hat{y}_i = \hat{y}_b + \text{Softmax}(QK^T)V.
\end{equation}
Finally, we obtained the final image $\hat{y}$ as a weighted average of the attention outputs. This design has three major drawbacks. First, it is memory intensive because the attention maps are computed on relatively high-resolution features (only downsampled by a factor of eight). Second, because each color term is processed independently, the module cannot model dependencies across color-specific adjustments. Third, the final weighted averaging can be brittle when color-term probabilities overlap at boundaries, often producing halo artifacts.
}

\rev{
\textbf{Proposed transformer-based fusion.}
Standard spatial self-attention is effective but suffers from quadratic time and memory complexity due to the key--query dot-product over spatial positions, which becomes prohibitive for high-resolution inputs.} To address this, we adopt the transposed attention mechanism introduced by Zamir et al.~\cite{zamir2022restormer}, which computes cross-covariance across channels instead of across spatial locations. This yields an attention map whose cost scales linearly with image size. Moreover, rather than partitioning the image into tokens (patches), we operate directly on dense feature maps, which is more suitable for low-level vision. Transposed attention has been widely adopted in restoration and enhancement tasks due to its favorable efficiency/accuracy trade-off and its ability to capture long-range dependencies~\cite{cai2023retinexformer, chen2024comparative, Serrano2025revisiting}. Figure~\ref{fig:transformer} shows the architecture of our fusion block.

The fusion block consists of a Multi-Head Transposed Attention module followed by a Feed-Forward Network. We first concatenate the six globally adjusted RGB images to form an 18-channel input $I \in \mathbb{R}^{H \times W \times 18}$. A $3 \times 3$ convolution extracts low-level features $F \in \mathbb{R}^{H \times W \times C}$. Following Zhang et al.~\cite{zhang2019self}, we compute the {\it query} ($Q$), {\it key} ($K$), and {\it value} ($V$) projections from $F$ via Layer Normalization, a $1 \times 1$ convolution, and a $3 \times 3$ depth-wise convolution for each branch, producing $Q, K, V \in \mathbb{R}^{H \times W \times C}$. To build the transposed attention map $A$, we swap the usual roles of $Q$ and $K$:
\begin{equation}
    A = \text{Softmax}(K^TQ).
\end{equation}
We then multiply $A$ by $V$ and refine the result with a depth-wise convolution. As in Zamir et al.~\cite{zamir2022restormer}, we split the channels into multiple heads, allowing the module to learn diverse attention patterns in parallel.

The output of the attention module is further refined by a Feed-Forward Network (FFN), which processes features independently at each spatial location while sharing parameters across the image. Specifically, we normalize the attention features $F_a$ with Layer Normalization and process them through two parallel branches, each composed of a point-wise convolution, a depth-wise convolution, and a final $1 \times 1$ convolution. This FFN enriches local interactions while preserving efficiency. Finally, we apply a $3 \times 3$ convolution to map the features $F_f$ to three channels, producing the final enhanced sRGB image. In all experiments, we set $C{=}18$, matching the number of input channels.

\subsection{Loss Function}\label{subsec:loss_function}
Our training loss function comprises three terms. The initial term calculates the $L_2$ loss between the standardized image $\hat{y_b}$ and the ground truth $y$, with a weighting factor $\alpha$. The subsequent two terms compute the $L_2$ and $SSIM$ losses between the output of our model $\hat{y}$ and $y$. The primary objective of the first term is to obtain a good enough standardized output by the backbone. The other two terms are designed to assess the fidelity of the final output. In our experiments, we set $\alpha$ to 0.5. This value was determined to yield optimal performance and allows the image $\hat{y}_b$ to represent the scene colors accurately. The training loss is defined as:
\begin{equation}
    L(\hat{y_b}, \hat{y}, y) = \alpha ||y - \hat{y_b}||_2 + ||y - \hat{y}||_2 + (1 - SSIM(y, \hat{y})).
    \label{eq:loss}
\end{equation}

\section{Experiments}\label{sec:experiments}
\subsection{Implementation Details}~We trained our model using Adam~\cite{kingma2014adam}, an initial learning rate 1e-4, reduced by 50\% every 50 epochs. We use horizontal flips for augmenting the training data. We train the models for a fixed 200 epochs for all the datasets: MIT5K~\cite{bychkovsky2011mit5k}, PPR10K~\cite{liang2021ppr10k}, MSEC~\cite{afifi2021msec}, and SICE~\cite{cai2018sice}.

\subsection{Quantitative Results}
\textbf{Image Retouching.} We compare our method with state-of-the-art image retouching methods using the MIT-Adobe-5K dataset \cite{bychkovsky2011mit5k} and the PPR10K dataset \cite{liang2021ppr10k}. MIT5K~\cite{bychkovsky2011mit5k} consists of 5,000 images captured independently using several DSLR cameras and retouched by five artists. Following \cite{moran2021curl, chen2018deep, li2023flexicurve, moran2020deeplpf, wang2019underexposed} we only use the Expert C retouched images as ground truth, and the pre-processing proposed by Ni et al.~\cite{ni2020towards}. Specifically, we split the image into 4,500 images for training and 500 images for testing. PPR10K~\cite{liang2021ppr10k} is a portrait photo retouching dataset with 11,616 high-quality images retouched by three experts independently. We use the official splits, dividing the images into 8,875 and 2,286 for training and testing, respectively. Following \cite{yang2022adaint}, we conducted the experiments on the 360p augmented version, in which every image pair has five extra input versions with different manual settings. As in \cite{zeng2020learning, liang2021ppr10k, yang2022adaint}, we evaluate our method on the three expert-retouched versions of PPR10K.

In Table~\ref{tab:MIT5K}, we report the PSNR, SSIM, LPIPS, $\Delta E_{00}$, and $\Delta E_{ab}$ of the state-of-the-art methods and our approach for the MIT5K dataset. We also report the inference time for a 480p image on an AMD EPYC 7642 and an NVIDIA A40. Our full architecture outperforms contemporary image-to-image, curve estimation, and LUT-based methods. Table \ref{tab:ppr10k} reports the PSNR and $\Delta E_{ab}$ on the PPR10K dataset. Our method consistently outperforms the previous state-of-the-art methods across the three experts.

\begin{table}[t!]
  \caption{Image retouching. Quantitative comparisons on the MIT-Adobe-5K dataset~\cite{bychkovsky2011mit5k}. - means the source code or models are unavailable, or results for the corresponding metric were not in the original paper.}
  \label{tab:MIT5K}
  \centering
  \setlength{\tabcolsep}{3.6pt}
  \begin{tabular}{lcccccc}
    \toprule
    \multicolumn{1}{c}{\centering Method} & PSNR & SSIM  & LPIPS & $\Delta E_{00}$ & $\Delta E_{ab}$ & Time (ms)\\
    \midrule
    UPE~\cite{wang2019underexposed} & 21.88 & 0.853 & - & - & 10.80 & - \\
    DPE~\cite{chen2018deep} & 23.75 & 0.908 & - & - & 9.34 & - \\
    DeepLPF~\cite{moran2020deeplpf} & 24.73 & 0.916 & 0.078 & 7.80 & 7.99 & 136\\
    3DLUT \cite{zeng2020learning} & 25.29 & 0.923 & 0.043 & 6.76 & 7.55 & 13 \\
    SepLUT \cite{yang2022seplut} & 25.47 & 0.921 & 0.042 & 6.71 & 7.49 & 10 \\
    AdaInt \cite{yang2022adaint} & 25.49 & 0.926 & 0.041 & 6.69 & 7.47 & 13 \\
    \rev{RSFNet \cite{ouyang2023rsfnet}} & \rev{25.49} & \rev{0.924} & \rev{0.043} & \rev{6.73} & \rev{7.23} & \rev{121}\\
    NamedCurves \cite{serrano2025namedcurves} & 25.59 & 0.936 & 0.038 & 6.07 & 7.40 & 26 \\
    BGLUT \cite{kim2025bilateral} & 25.66 & 0.930 & 0.038 & 6.06 & 7.29 & 15 \\
    NamedCurves+ & \textbf{25.75} & \textbf{0.940} & \textbf{0.035} & \textbf{6.01} & \textbf{7.27} & 19 \\    
  \bottomrule
  \end{tabular}
  \vspace{-3mm}
\end{table}

\begin{table}[t!]
    \caption{Image retouching. Quantitative comparisons on PPR10K.}
    \label{tab:ppr10k}
    \centering
    \setlength{\tabcolsep}{3.6pt}
    \begin{tabular}{lcccccc}
    \toprule
    \multicolumn{1}{c}{} & \multicolumn{2}{c}{ Expert A } & \multicolumn{2}{c}{ Expert B } & \multicolumn{2}{c}{  Expert C } \\
    \cmidrule{2-3}\cmidrule(lr){4-5}\cmidrule{6-7}
    \multicolumn{1}{c}{\centering Method} & PSNR & $\Delta E_{ab}$ & PSNR & $\Delta E_{ab}$ 
 & PSNR & $\Delta E_{ab}$ \\
    \midrule
    HDRNet \cite{li2019hdrnet}              & 23.93 & 8.70 & 23.96 & 8.84 & 24.08 & 8.87 \\
    3DLUT \cite{zeng2020learning}           & 25.64 & 6.97 & 24.70 & 7.71 & 25.18 & 7.58 \\
    SepLUT \cite{yang2022seplut}            & 26.28 & 6.59 & 25.23 & 7.49 & 25.59 & 7.51 \\
    AdaInt \cite{yang2022adaint}            & 26.33 & 6.56 & 25.40 & 7.33 & 25.68 & 7.31 \\
    \rev{RSFNet \cite{ouyang2023rsfnet}} & \rev{25.58} & \rev{6.92} & \rev{24.81} & \rev{7.58} & \rev{25.52} & \rev{7.47} \\
    BGLUT \cite{kim2025bilateral} & 26.45 & 6.51 & 25.48 & 7.19 & 25.72 & 7.28 \\
    NamedCurves \cite{serrano2025namedcurves} & 26.81 & 6.48 & 25.91 & 7.18 & 25.69 & 7.27 \\
    NamedCurves+ & \textbf{26.85} & \textbf{6.45} & \textbf{25.95} & \textbf{7.14} & \textbf{25.77} & \textbf{7.24} \\
    \bottomrule
    \end{tabular}
\end{table}

\textbf{Tone Mapping.} We compare our method with state-of-the-art tone mapping methods using the MIT-Adobe-5K dataset \cite{bychkovsky2011mit5k}. In contrast with image retouching, the input images are in 16-bit CIE XYZ format and the outputs are sRGB images. Following Kim et al.~\cite{kim2025bilateral}, in Table \ref{tab:tone_mapping} we report PSNR, SSIM, and $\Delta E_{ab}$ of the previous state-of-the-art methods and our approach. We outperform all the current methods across all the metrics, including our previous work NamedCurves~\cite{serrano2025namedcurves}.

\begin{table}[t!]
  \caption{Tone mapping. Quantitative comparisons on MIT-Adobe-5K}
  \label{tab:tone_mapping}
  \centering
  \begin{tabular}{lccc}
    \toprule
    \multicolumn{1}{c}{\centering Method} & PSNR $\uparrow$ & SSIM $\uparrow$ & $\Delta E_{ab}$ $\downarrow$\\
    \midrule
    UPE~\cite{wang2019underexposed} & 21.56 & 0.837 & 12.29 \\
    DPE~\cite{chen2018deep} & 22.93 & 0.894 & 11.09 \\
    3DLUT \cite{zeng2020learning} & 25.07 & 0.920 & 7.55 \\
    SepLUT \cite{yang2022seplut} & 25.43 & 0.922 & 7.43 \\
    AdaInt \cite{yang2022adaint} & 25.28 & 0.925 & 7.48 \\
    BGLUT \cite{kim2025bilateral} & 25.59 & 0.932 & 7.14 \\
    NamedCurves \cite{serrano2025namedcurves} & 25.60 & 0.932 & 7.15 \\
    NamedCurves+ (ours) & \textbf{25.66} & \textbf{0.933} & \textbf{7.12} \\
  \bottomrule
  \end{tabular}
    \vspace{-2mm}
\end{table}

\textbf{Exposure Correction.} We evaluate our approach against state-of-the-art methods and NamedCurves~\cite{serrano2025namedcurves} on the ME~\cite{afifi2021msec} and SICE~\cite{cai2018sice} datasets. The ME dataset contains over 24,000 images with the widest range of exposure values, paired with corresponding properly exposed reference images. This dataset was generated from the MIT-5K dataset using Adobe Camera RAW, which simulates different exposure values as a camera would apply them. Adobe Camera RAW uses metadata to ensure accurate emulation of nonlinear camera rendering procedures. The ME dataset is divided into 17,675 training images, 750 validation images, and 5,905 testing images. The SICE dataset, on the other hand, comprises 589 high-resolution multi-exposure sequences, totaling 4,413 images captured by seven types of consumer-grade cameras. Following Huang et al.~\cite{huang2022fecnet}, in Table~\ref{tab:exposure}, we report the PSNR and SSIM for underexposed images, overexposed images, and the average of all testing images. Our approach achieves the best PSNR and SSIM on both datasets, except for PSNR on overexposed images in the ME dataset and SSIM on overexposed images in the SICE dataset, where our method ranks second. \rev{These quantitative results highlight the effectiveness of our two-step approach. First, the backbone network recovers lost details, followed by the Bezier tone curves and transformer block, which enhance the image’s color and overall quality.}

\begin{table*}[t!]
  \caption{Exposure correction. Quantitative comparisons on the MSEC and SICE datasets.}
  \footnotesize
  \label{tab:exposure}
  \centering
  \setlength{\tabcolsep}{6pt}
  \begin{tabular}{lcccccccccccc}
    \toprule
    & \multicolumn{6}{c}{\centering MSEC~\cite{afifi2021msec}} & \multicolumn{6}{c}{\centering SICE~\cite{cai2018sice}} \\
    \cmidrule(lr){2-7} \cmidrule(lr){8-13}
    & \multicolumn{2}{c}{\centering Under} & \multicolumn{2}{c}{\centering Over} & \multicolumn{2}{c}{\centering Average} & \multicolumn{2}{c}{\centering Under} & \multicolumn{2}{c}{\centering Over} & \multicolumn{2}{c}{\centering Average} \\
    \cmidrule(l){2-3}\cmidrule(lr){4-5}\cmidrule(r){6-7} \cmidrule(l){8-9}\cmidrule(lr){10-11}\cmidrule(r){12-13}
    \multicolumn{1}{c}{\centering Method} & PSNR & SSIM & PSNR & SSIM & PSNR & SSIM & PSNR & SSIM & PSNR & SSIM & PSNR & SSIM \\
    \midrule
    CLAHE~\cite{zuiderveld1994clahe} & 16.77 & 0.6211 & 14.45 & 0.5842 & 15.38 & 0.5990 & 12.69 & 0.5037 & 10.21 & 0.4878 & 11.45 & 0.4942 \\
    RetinexNet~\cite{Chen2018Retinexnet} & 12.13 & 0.6209 & 10.47 & 0.5953 & 11.14 & 0.6048 & 12.94 & 0.5171 & 12.87 & 0.5252 & 12.90 & 0.5212 \\
    ZeroDCE~\cite{guo2020zero} & 14.55 & 0.5887 & 10.40 & 0.5142 & 12.06 & 0.5441 & 16.92 & 0.6330 & 7.10 & 0.4292 & 12.02 & 0.5311 \\
    MSEC~\cite{afifi2021msec} & 20.52 & 0.8129 & 19.79 & 0.8156 & 20.35 & 0.8210 & 19.62 & 0.6512 & 17.59 & 0.6560 & 18.58 & 0.6536 \\
    CMEC~\cite{nsampi2021cmec} & 22.23 & 0.8140 & 22.75 & 0.8336 & 22.54 & 0.8257 & 17.68 & 0.6592 & 18.17 & 0.6811 & 17.93 & 0.6702 \\
    I-SID~\cite{huang2022exposure} & 22.59 & 0.8423 & 22.36 & 0.8519 & 22.45 & 0.8481 & 21.30 & 0.6645 & 19.63 & 0.6941 & 20.47 & 0.6793 \\
    I-DRBN-4~\cite{huang2022exposure} & 22.72 & 0.8544 & 22.11 & 0.8521 & 22.35 & 0.8530 & 21.77 & 0.7052 & 19.57 & \textbf{0.7267} & 20.67 & 0.7160 \\
    \rev{RSFNet~\cite{ouyang2023rsfnet}} & \rev{22.61} & \rev{0.8457} & \rev{22.16} & \rev{0.8443} & \rev{22.39} & \rev{0.8450} & \rev{21.42} & \rev{0.6724} & \rev{19.03} & \rev{0.6892} & \rev{20.23} & \rev{0.6808}\\
    FECNet~\cite{huang2022fecnet} & 22.96 & 0.8598 & \textbf{23.22} & 0.8748 & 23.12 & 0.8688 & 22.01 & 0.6737 & 19.91 & 0.6961 & 20.96 & 0.6849 \\
    NamedCurves~\cite{serrano2025namedcurves} & 23.11 & 0.8698 & 22.98 & 0.8771 & 23.06 & 0.8726 & 21.89 & 0.7154 & 21.95 & 0.7142 & 21.92 & 0.7148 \\
    NamedCurves+ (ours) & \textbf{23.24} & \textbf{0.8710} & 23.09 & \textbf{0.8839} & \textbf{23.15} & \textbf{0.8780} & \textbf{22.07} & \textbf{0.7194} & \textbf{22.10} & 0.7194 & \textbf{22.08} & \textbf{0.7194}\\
    \bottomrule
  \end{tabular}
  \vspace{-2mm}
\end{table*}

\begin{table}[t!]
  \caption{\rev{Module contribution ablation study on MIT-Adobe-5K(retouching) and MSEC (exposure correction).}}
  \label{tab:module_ablation}
  \centering
  \color{myblack}
  \setlength{\tabcolsep}{3.6pt}
  \begin{tabular}{lcccccc}
    \toprule
    & \multicolumn{4}{c}{MIT-5K} & \multicolumn{2}{c}{MSEC} \\
    \cmidrule(r){2-5}     \cmidrule(l){6-7}
    \multicolumn{1}{c}{\centering Method} & PSNR & SSIM & LPIPS & $\Delta E_{00}$ & PSNR & SSIM \\
    \midrule
    w/o Backbone & 24.30 & 0.916 & 0.052 & 7.16 & 20.98 & 0.848 \\
    w/o naming & 25.13 & 0.922 & 0.045 & 6.55 & 22.36 & 0.861\\
    w/o Transformer & 25.41 & 0.929 & 0.041 & 6.47 & 22.72 & 0.863\\  
    NamedCurves & 25.59 & 0.936 & 0.038 & 6.07 & 23.06 & 0.873 \\
    NamedCurves+ & 25.75 & 0.940 & 0.035 & 6.01 & 23.15 & 0.878\\
  \bottomrule
  \end{tabular}
\end{table}

\subsection{Qualitative Results}\label{subsec:qualitative_results}

\begin{figure*}[t!]
    \centering
    \includegraphics[width=.19\linewidth]{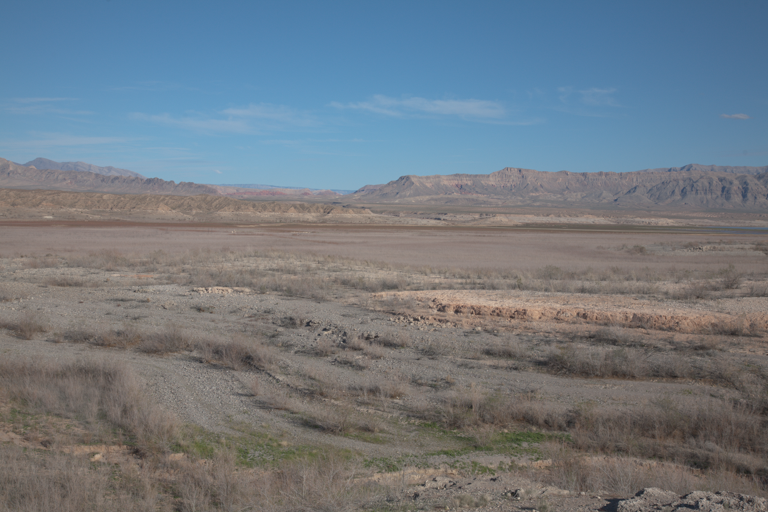}
    \includegraphics[width=.19\linewidth]{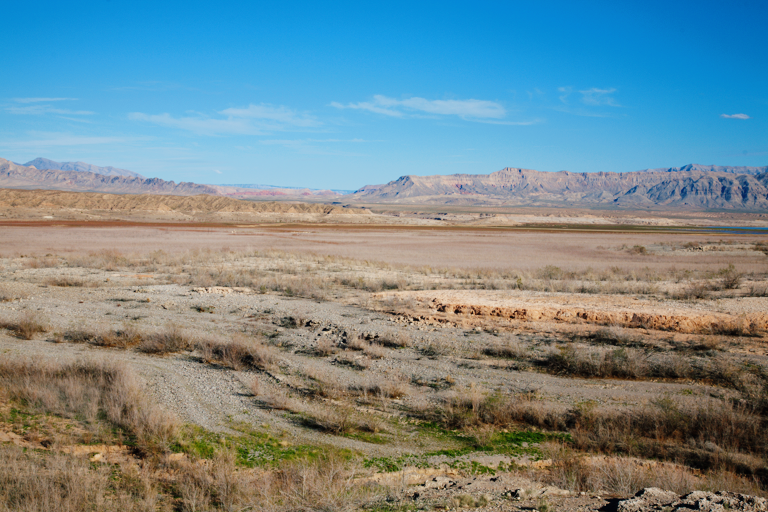}
    \includegraphics[width=.19\linewidth]{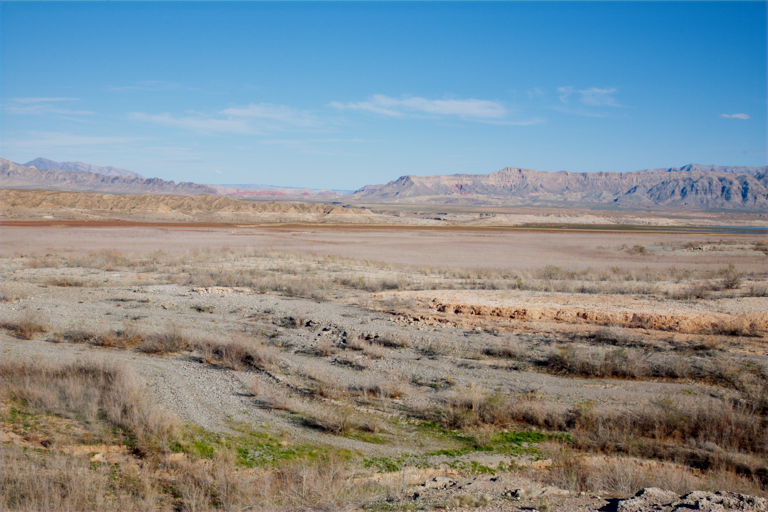}
    \includegraphics[width=.19\linewidth]{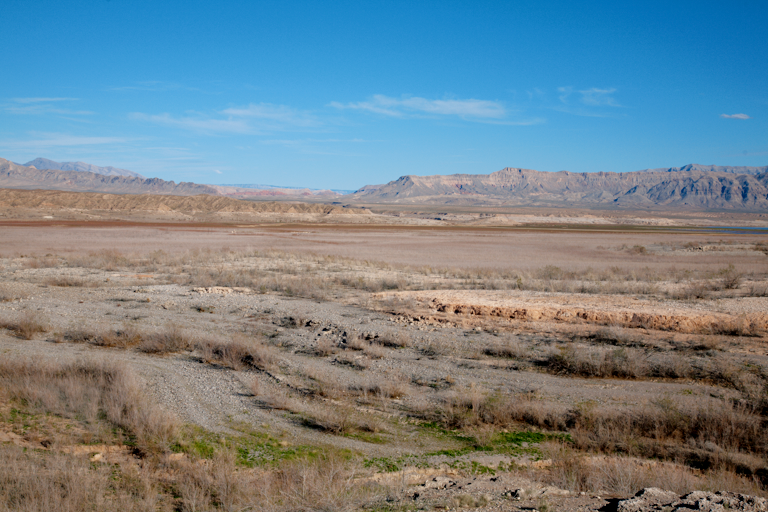}
    \includegraphics[width=.19\linewidth]{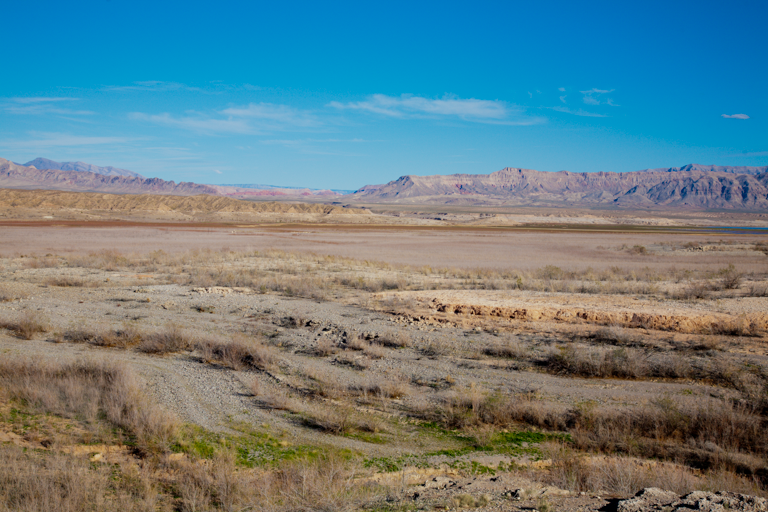} \\
    \vspace{1mm}
    \includegraphics[width=.19\linewidth]{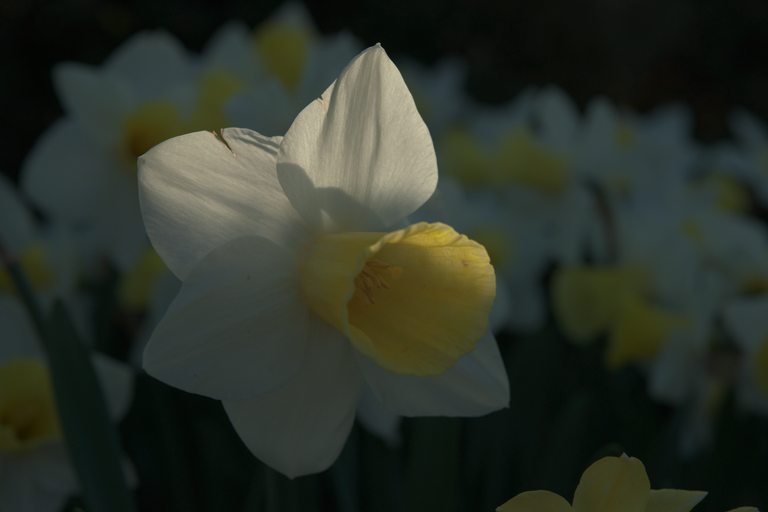}
    \includegraphics[width=.19\linewidth]{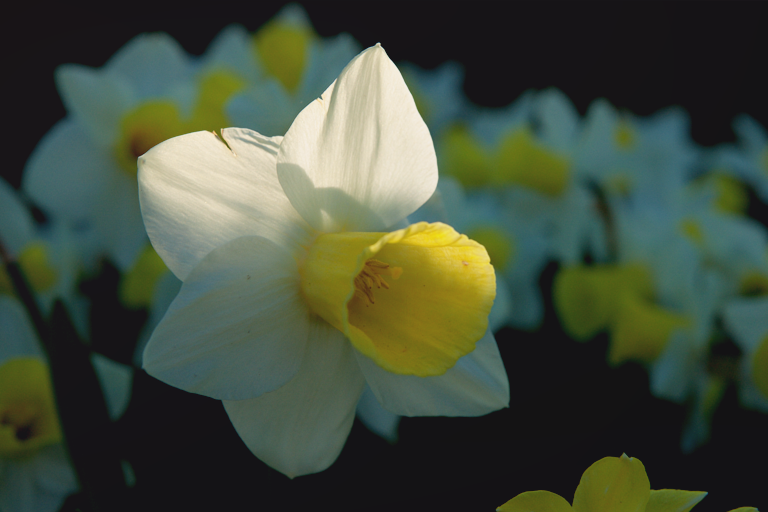}
    \includegraphics[width=.19\linewidth]{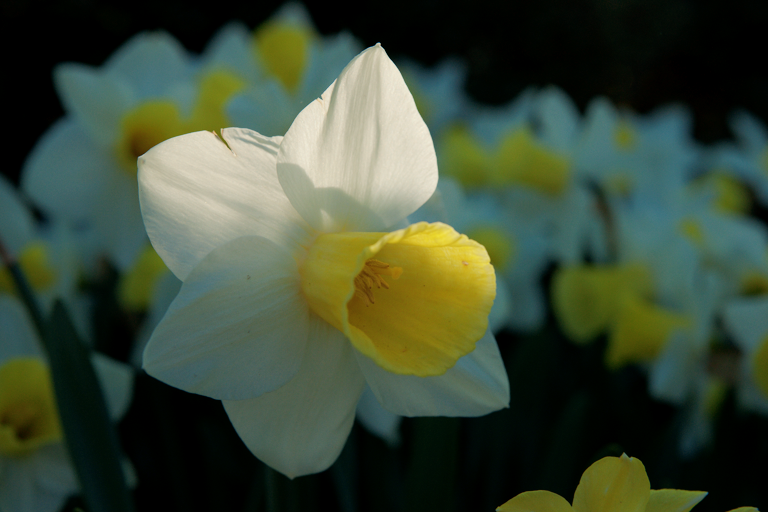}
    \includegraphics[width=.19\linewidth]{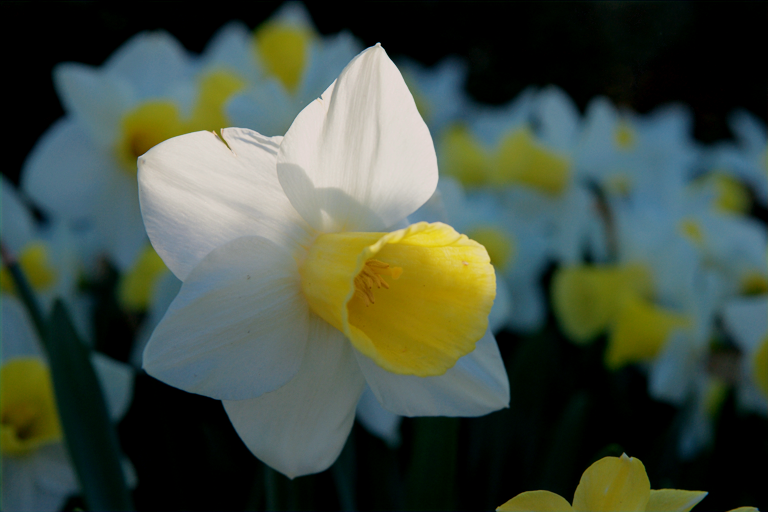}
    \includegraphics[width=.19\linewidth]{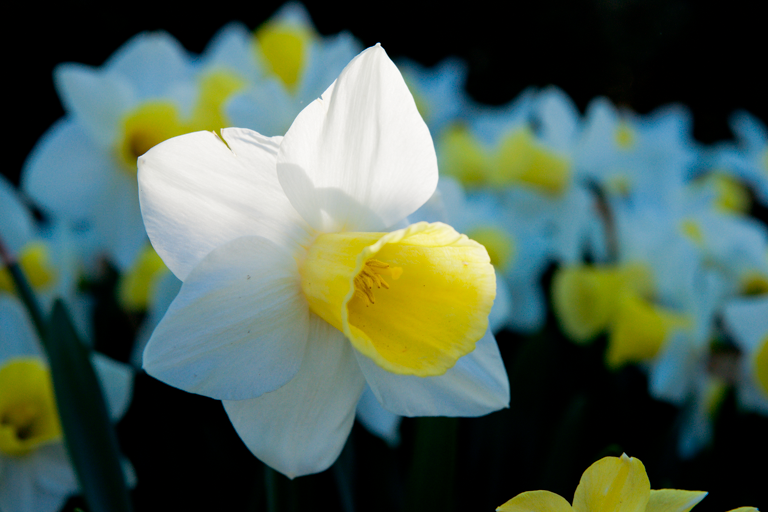}\\
    \makebox[0.19\textwidth]{\small Input}
    \makebox[0.19\textwidth]{\small BGLUT~\cite{kim2025bilateral}}
    \makebox[0.19\textwidth]{\small NamedCurves~\cite{serrano2025namedcurves}}
    \makebox[0.19\textwidth]{\small Ours}
    \makebox[0.19\textwidth]{\small Ground truth}
    \makebox[\textwidth]{\small (a) MIT5K dataset qualitative results.}

    \vspace{4mm}

    \includegraphics[width=.19\linewidth]{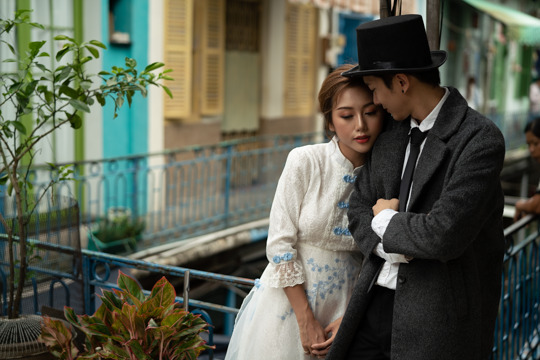}
    \includegraphics[width=.19\linewidth]{images/qualitative/ppr10k/1365_7.jpg}
    \includegraphics[width=.19\linewidth]{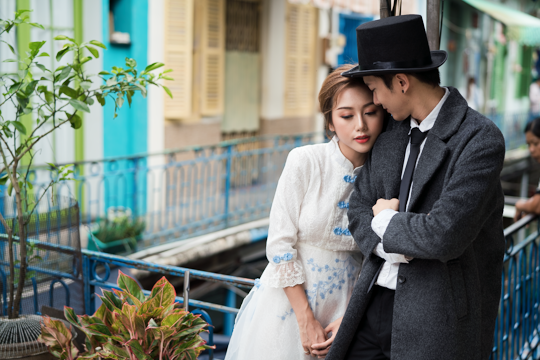}
    \includegraphics[width=.19\linewidth]{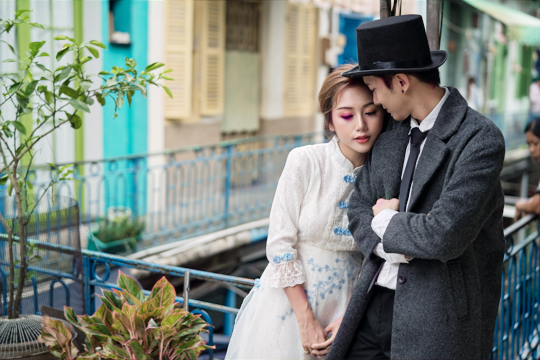}
    \vspace{1mm}
    \includegraphics[width=.19\linewidth]{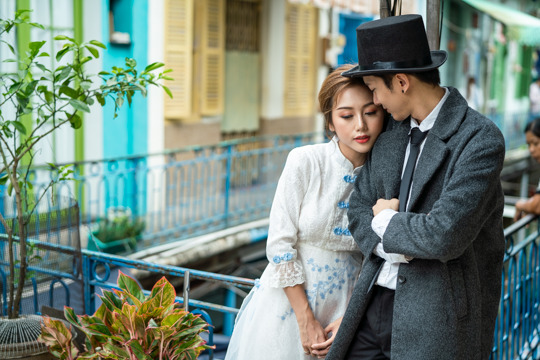} \\
    \includegraphics[width=.19\linewidth]{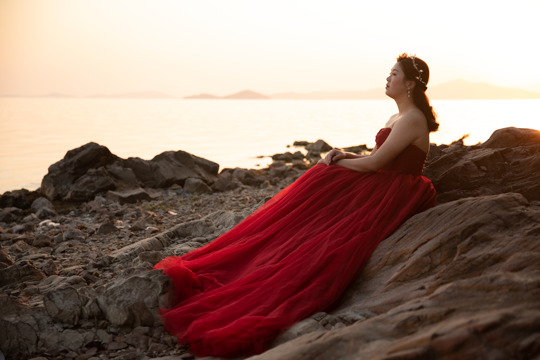}
    \includegraphics[width=.19\linewidth]{images/qualitative/ppr10k/1376_3.jpg}
    \includegraphics[width=.19\linewidth]{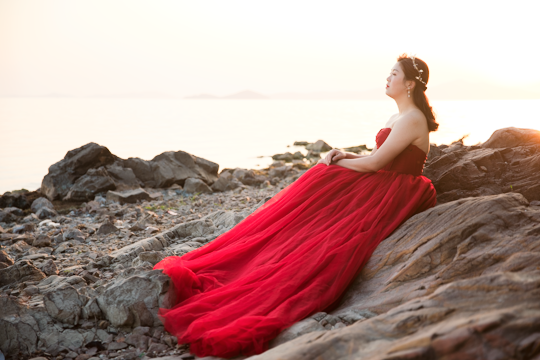}
    \includegraphics[width=.19\linewidth]{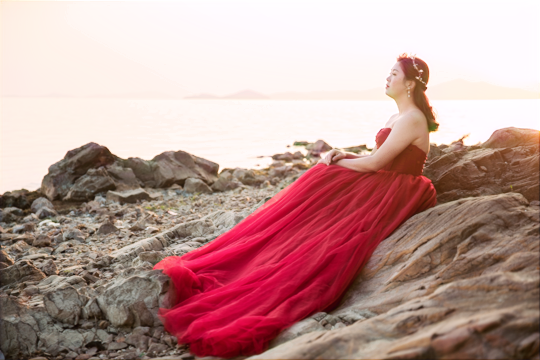}
    \includegraphics[width=.19\linewidth]{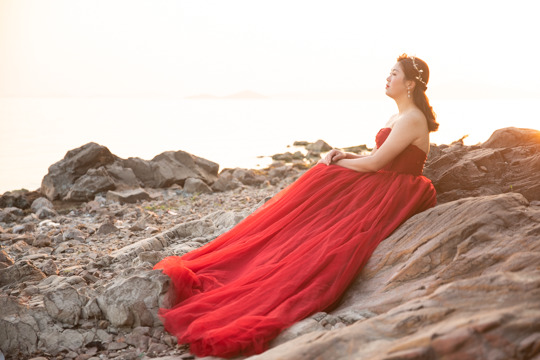}\\
    \makebox[0.19\textwidth]{\small Input}
    \makebox[0.19\textwidth]{\small BGLUT~\cite{kim2025bilateral}}
    \makebox[0.19\textwidth]{\small NamedCurves~\cite{serrano2025namedcurves}}
    \makebox[0.19\textwidth]{\small Ours}
    \makebox[0.19\textwidth]{\small Ground truth}
    \makebox[\textwidth]{\small (b) PPR10K dataset qualitative results.}
    \caption{Qualitative results on image-retouching: (a) MIT5K, and (b) PPR10K datasets.}
    \vspace{-2mm}
  \label{fig:qualitative1}
\end{figure*}

\begin{figure*}[t!]
    \centering
    \includegraphics[width=.19\linewidth]{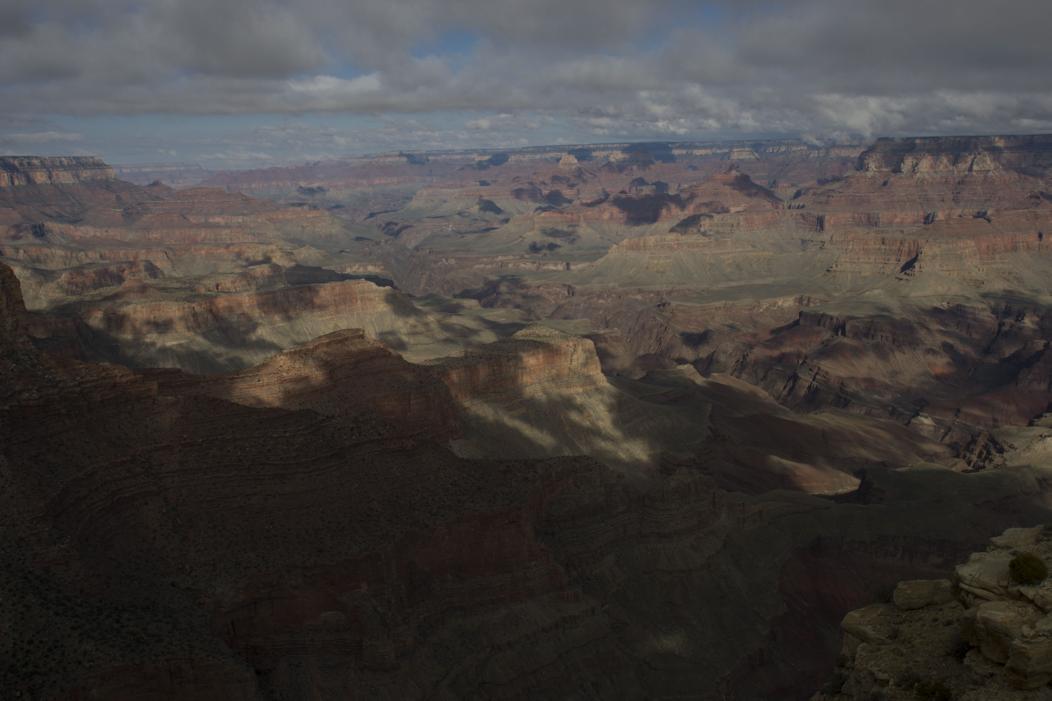}
    \includegraphics[width=.19\linewidth]{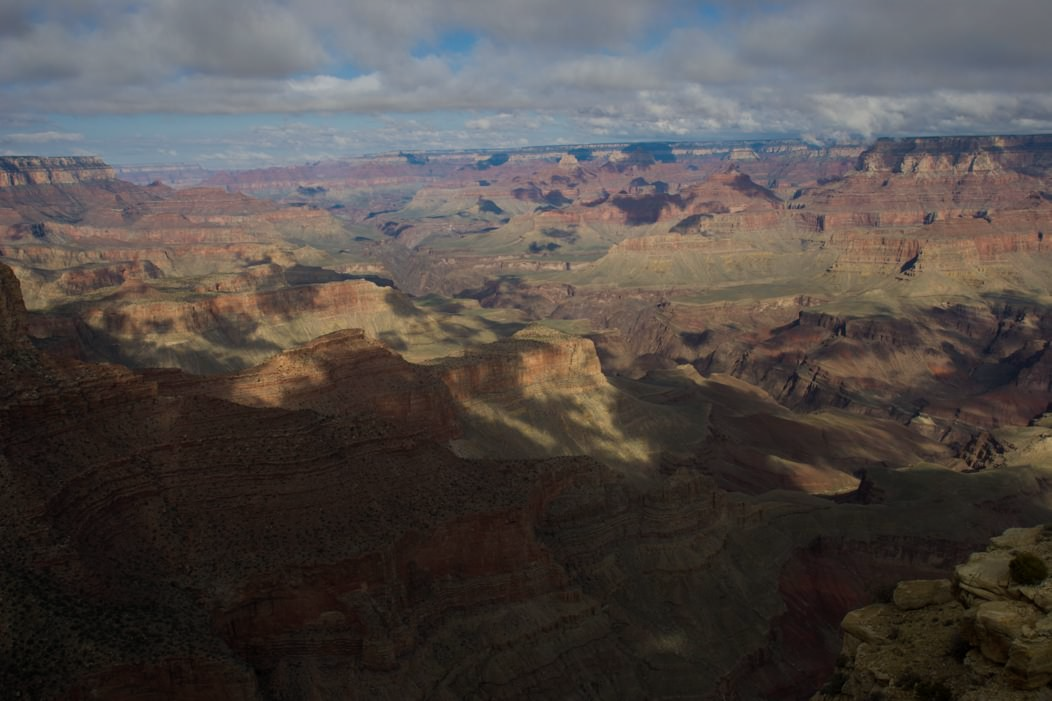}
    \includegraphics[width=.19\linewidth]{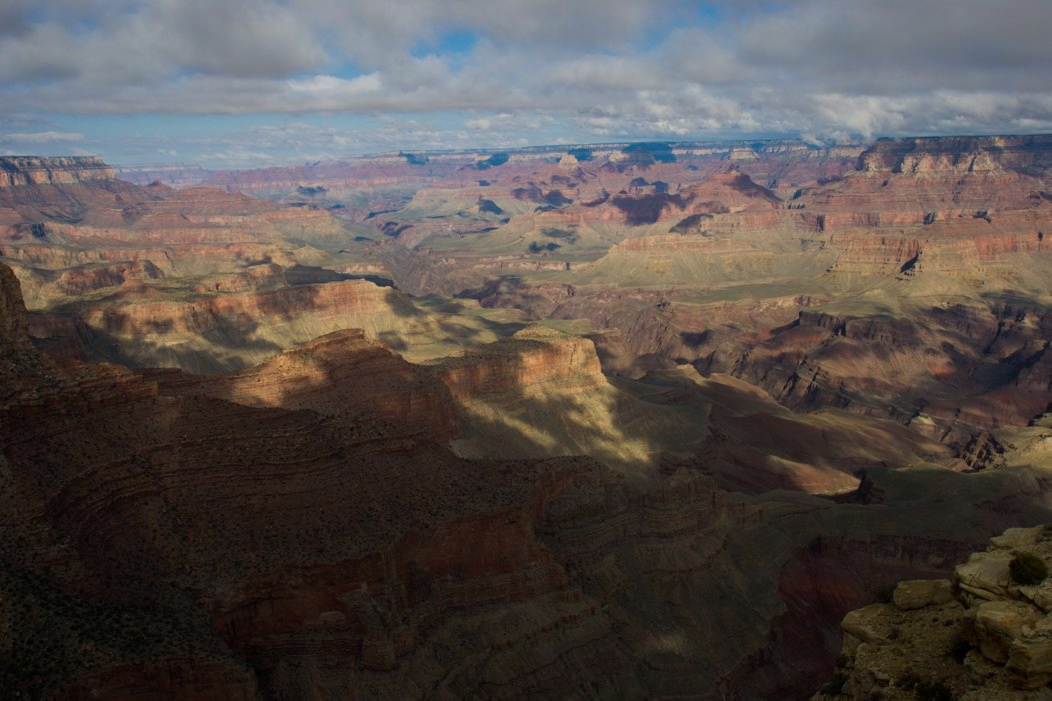}
    \includegraphics[width=.19\linewidth]{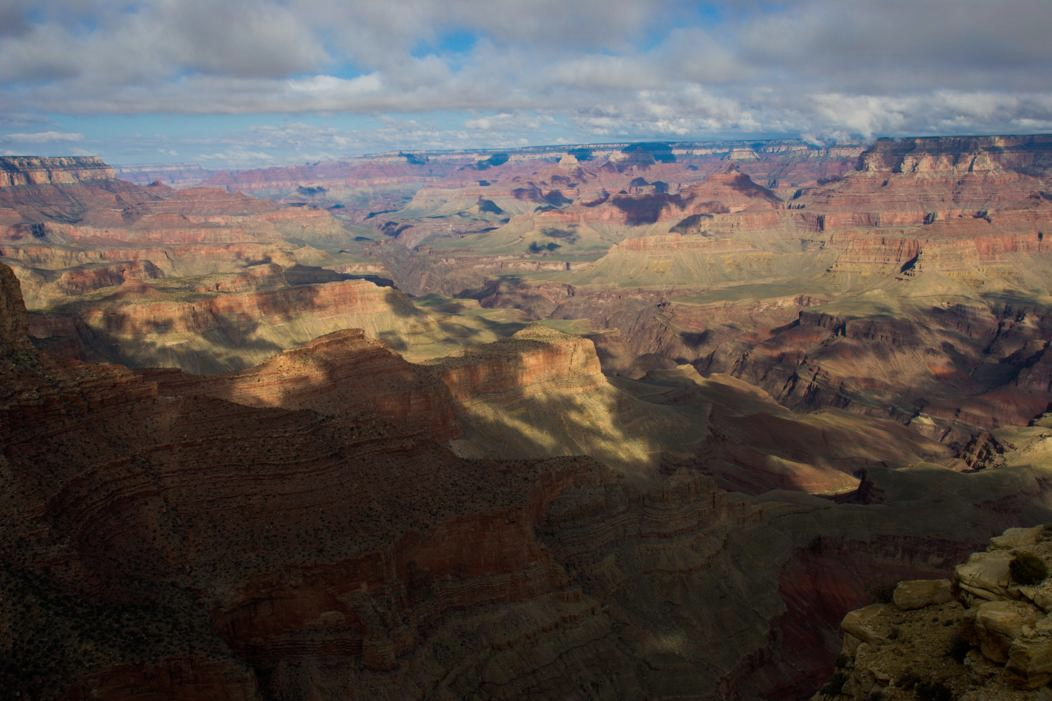}
    \includegraphics[width=.19\linewidth]{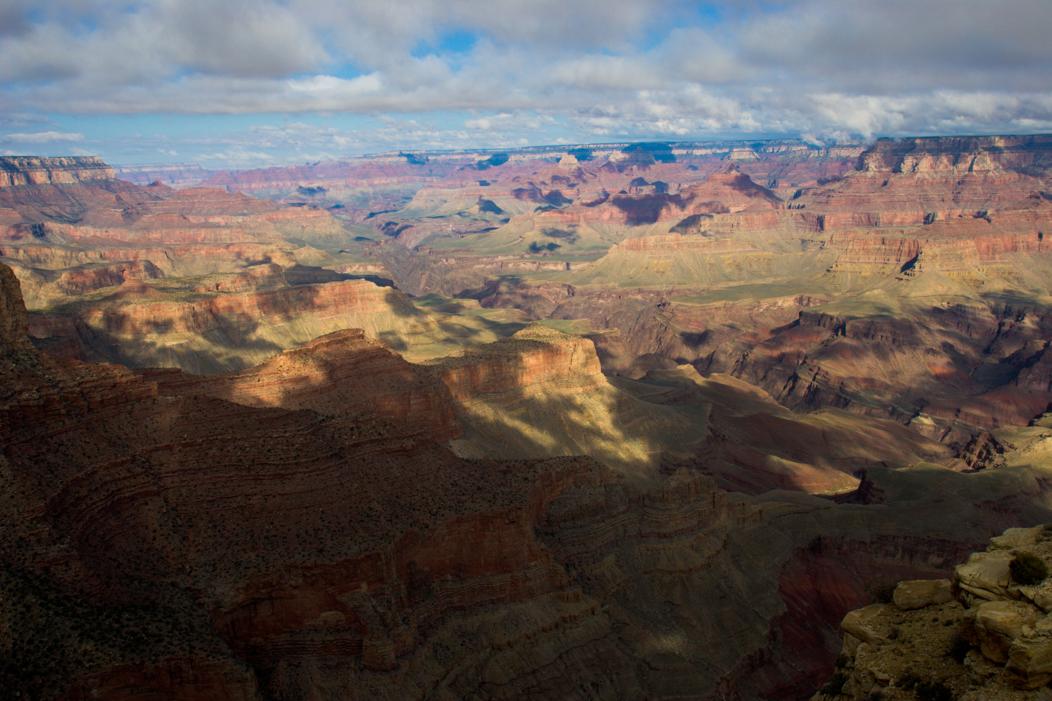} \\
    \vspace{1mm}
    \includegraphics[width=.19\linewidth]{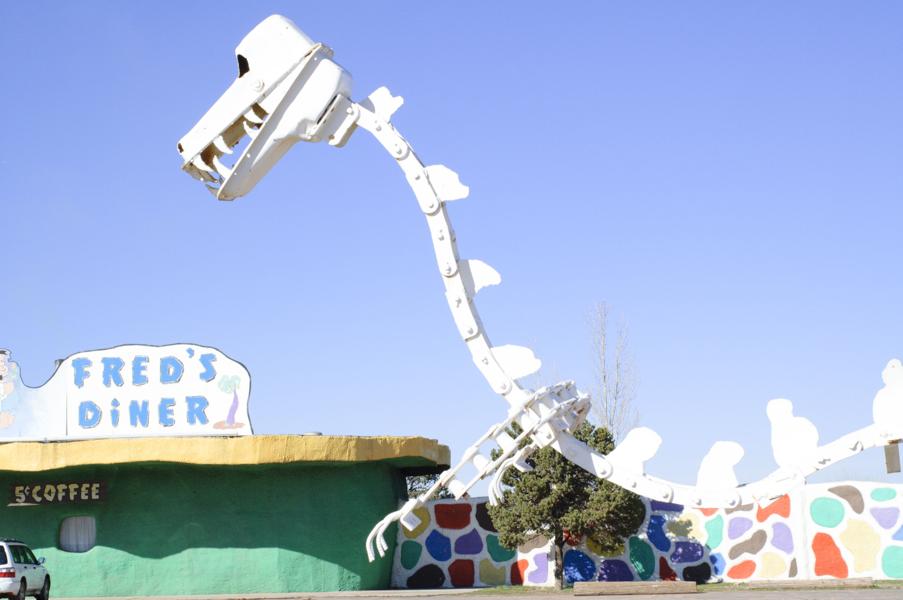}
    \includegraphics[width=.19\linewidth]{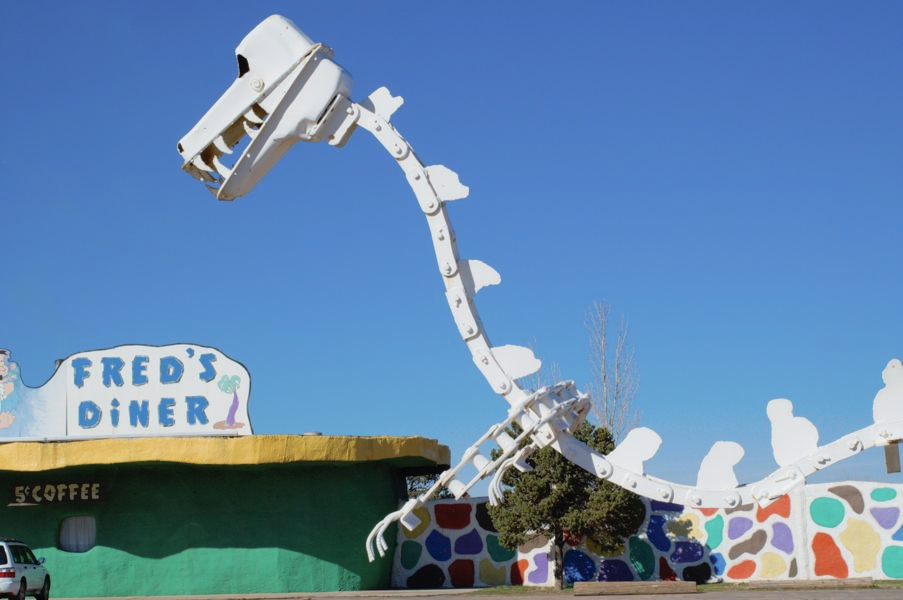}
    \includegraphics[width=.19\linewidth]{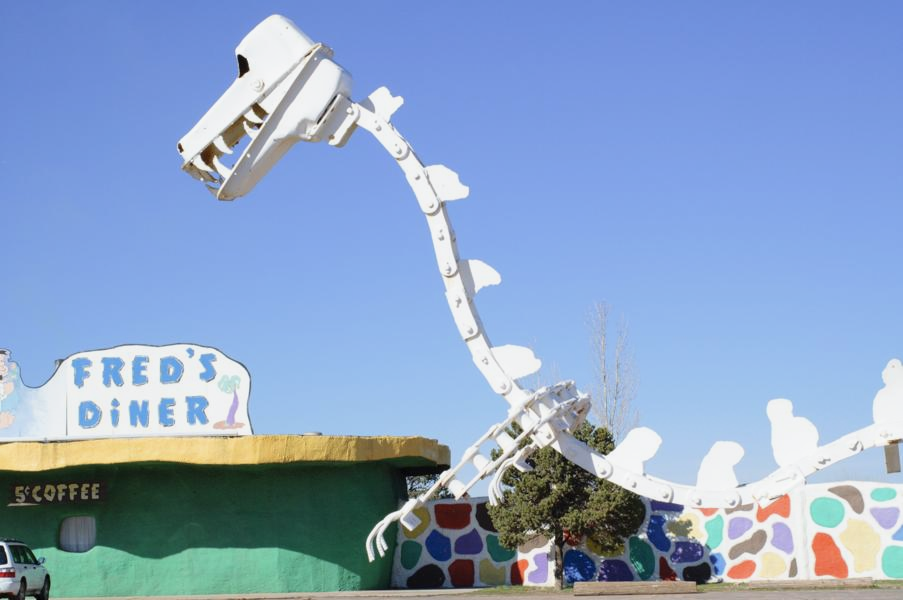}
    \includegraphics[width=.19\linewidth]{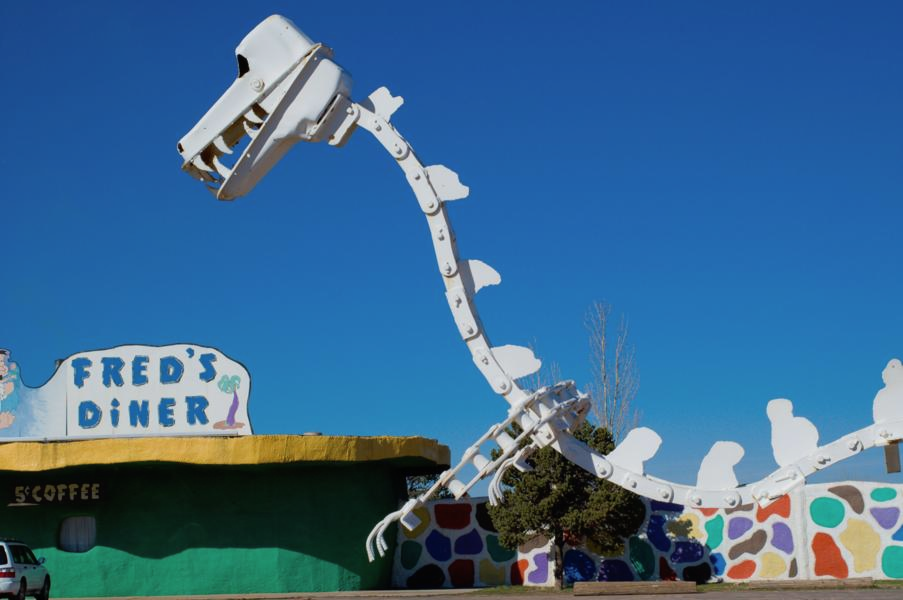}
    \includegraphics[width=.19\linewidth]{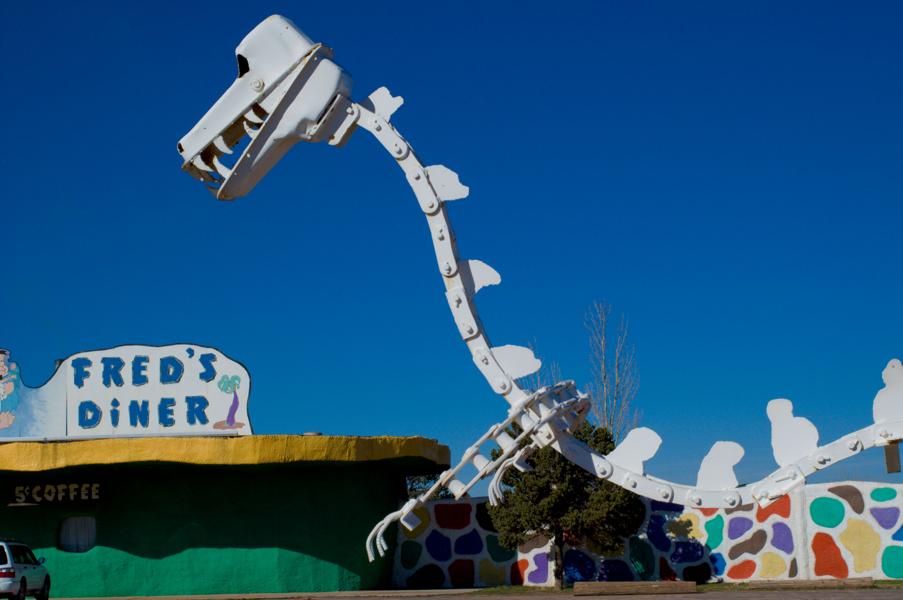}\\
    \makebox[0.19\textwidth]{\small Input}
    \makebox[0.19\textwidth]{\small FECNet~\cite{huang2022fecnet}}
    \makebox[0.19\textwidth]{\small NamedCurves~\cite{serrano2025namedcurves}}
    \makebox[0.19\textwidth]{\small Ours}
    \makebox[0.19\textwidth]{\small Ground truth}
    \makebox[\textwidth]{\small (a) MSEC dataset qualitative results.}
    \vspace{1mm}

    \includegraphics[width=.19\linewidth]{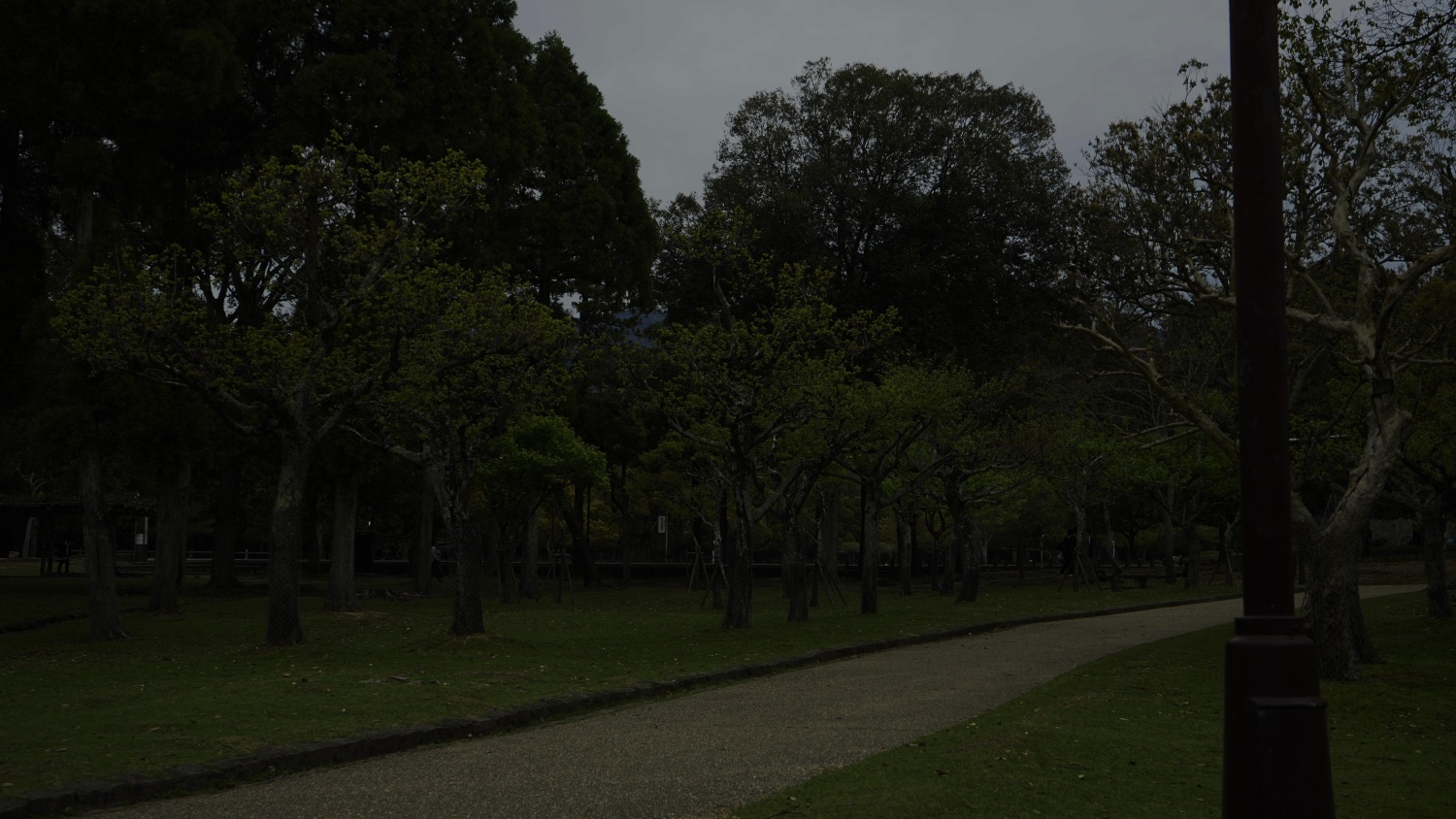}
    \includegraphics[width=.19\linewidth]{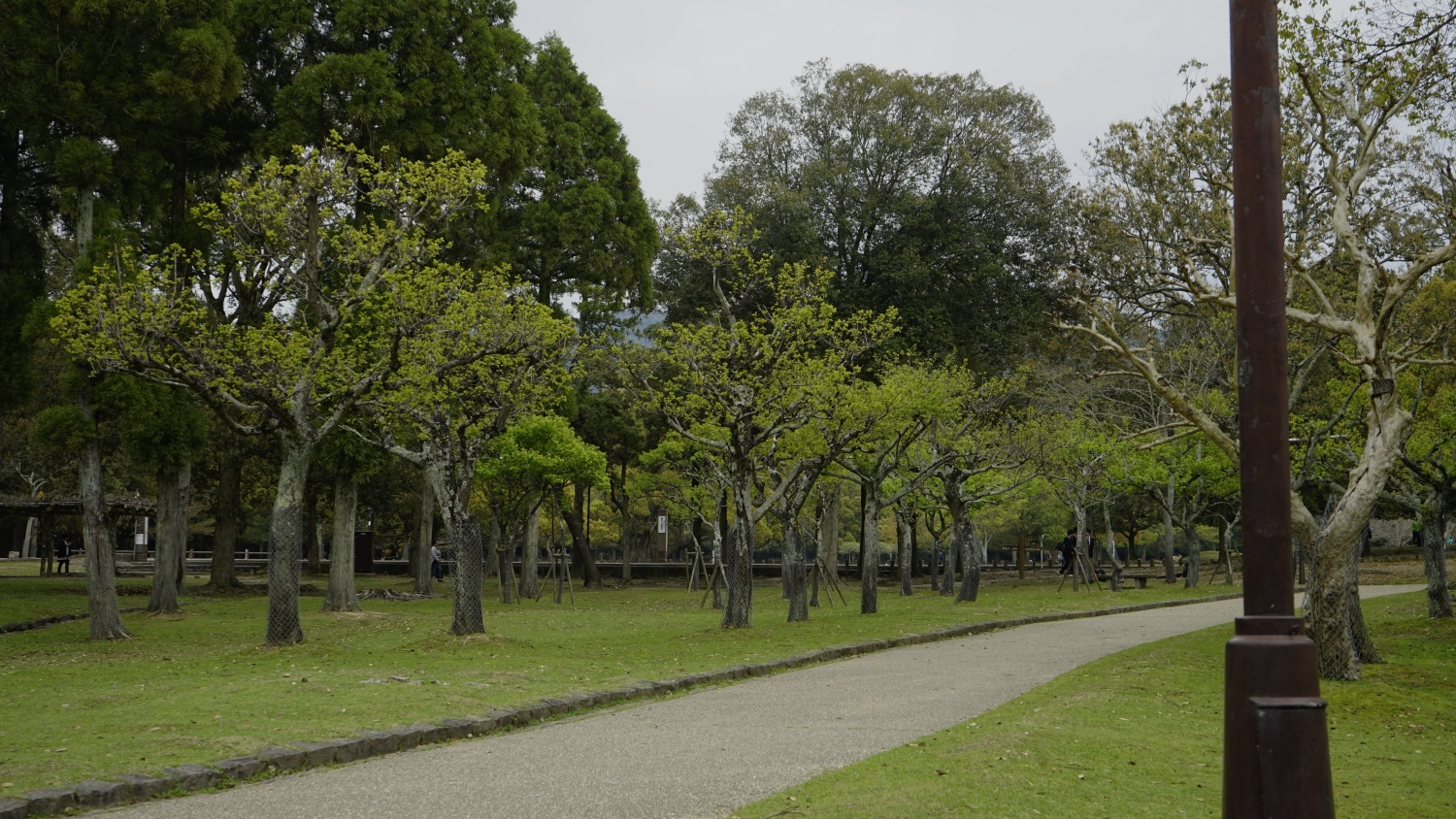}
    \includegraphics[width=.19\linewidth]{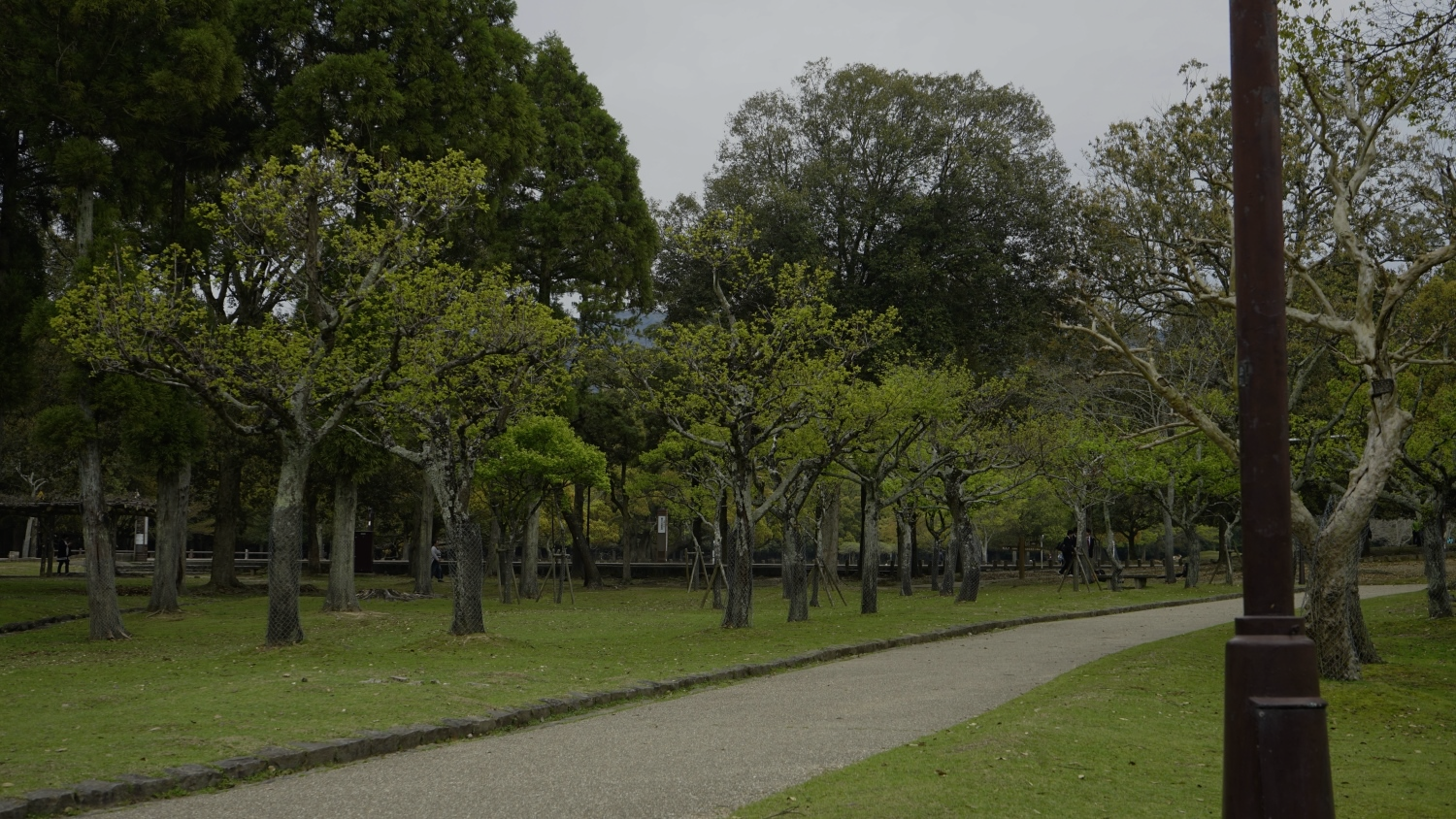}
    \includegraphics[width=.19\linewidth]{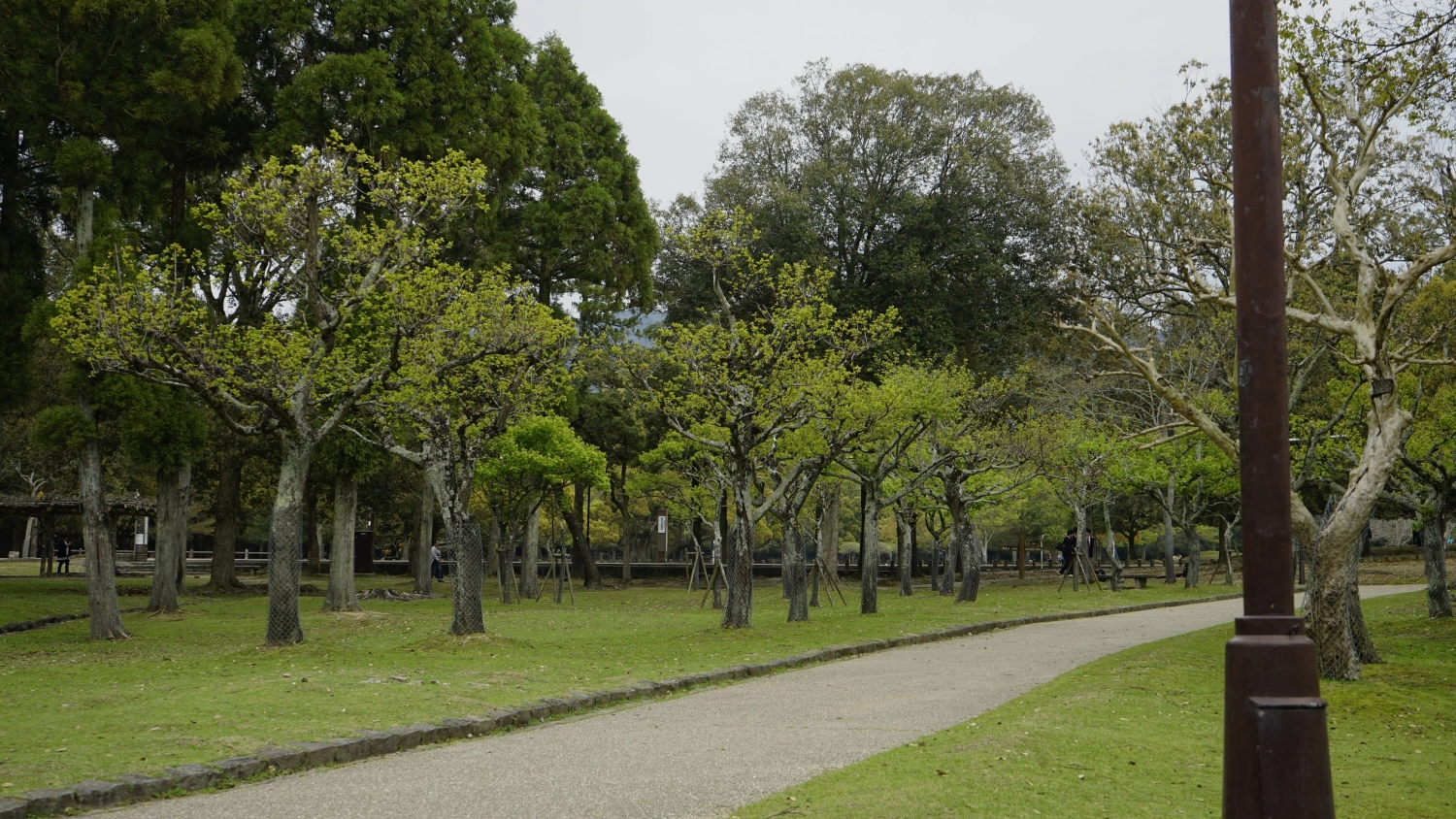}
    \includegraphics[width=.19\linewidth]{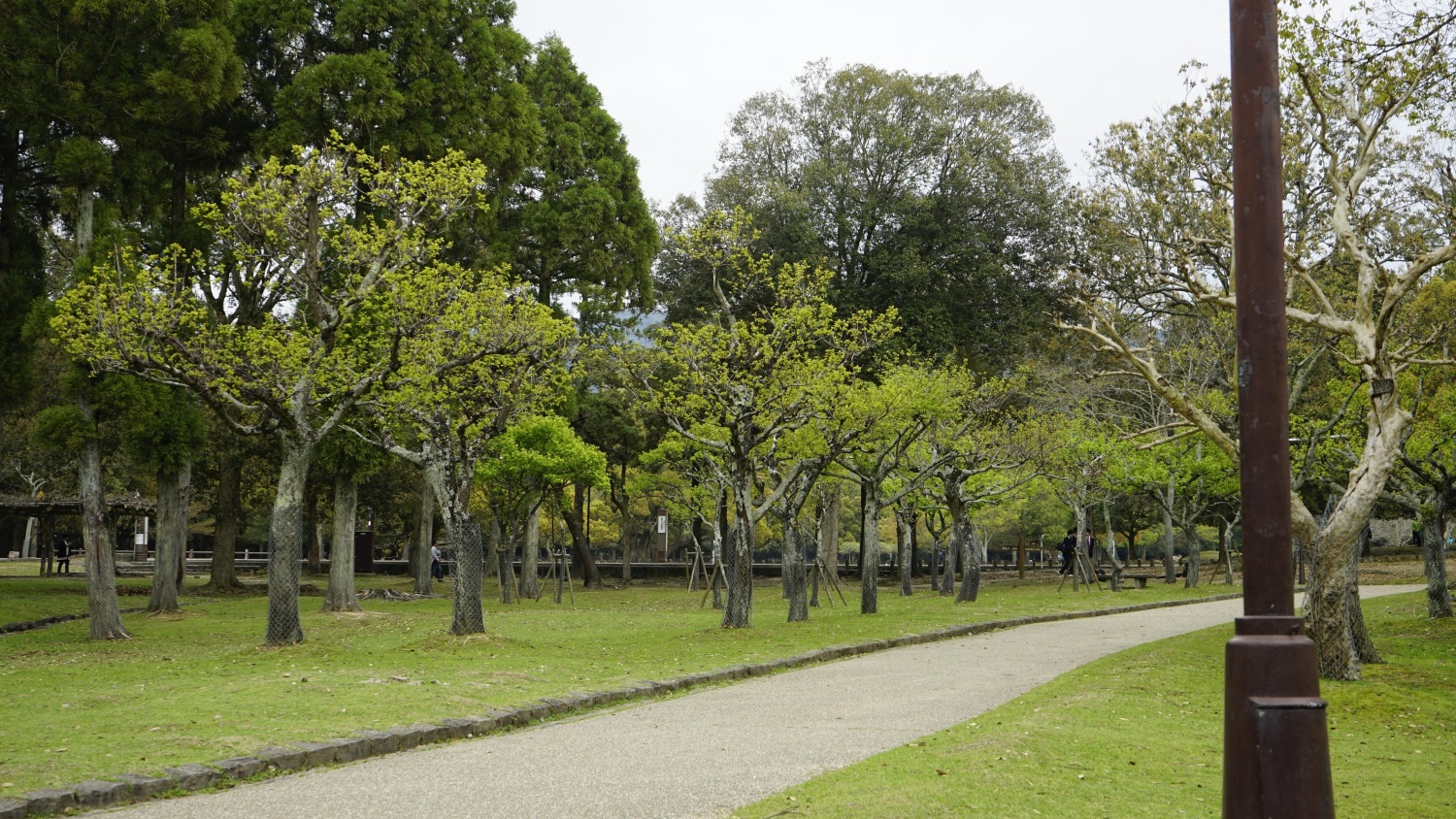} \\
    \vspace{1mm}
    \includegraphics[width=.19\linewidth]{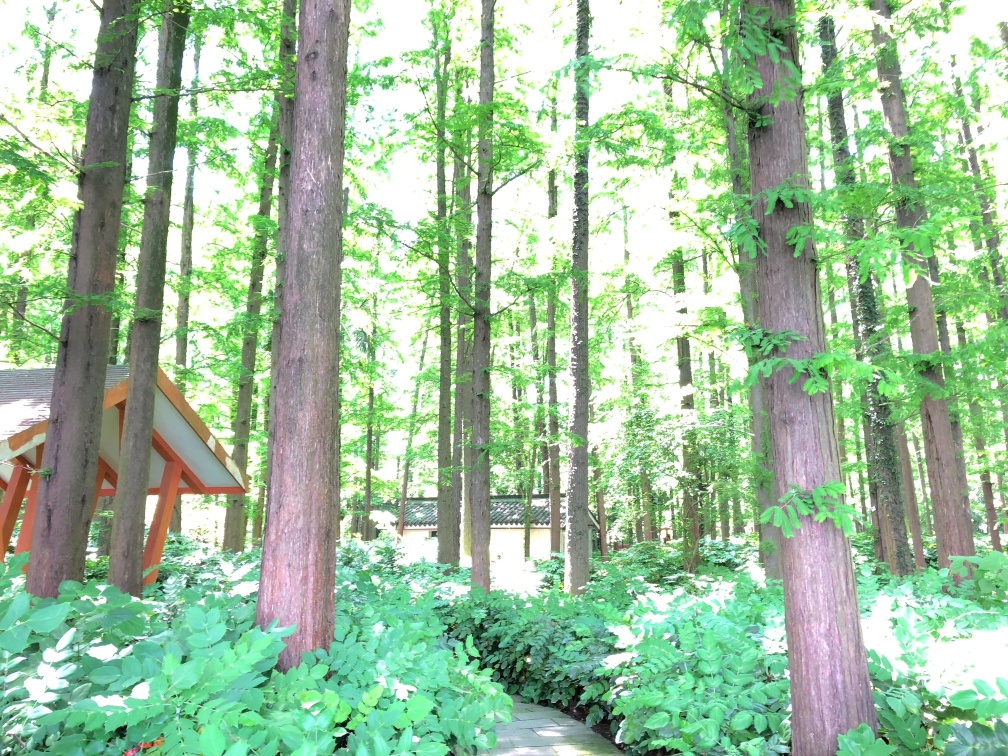}
    \includegraphics[width=.19\linewidth]{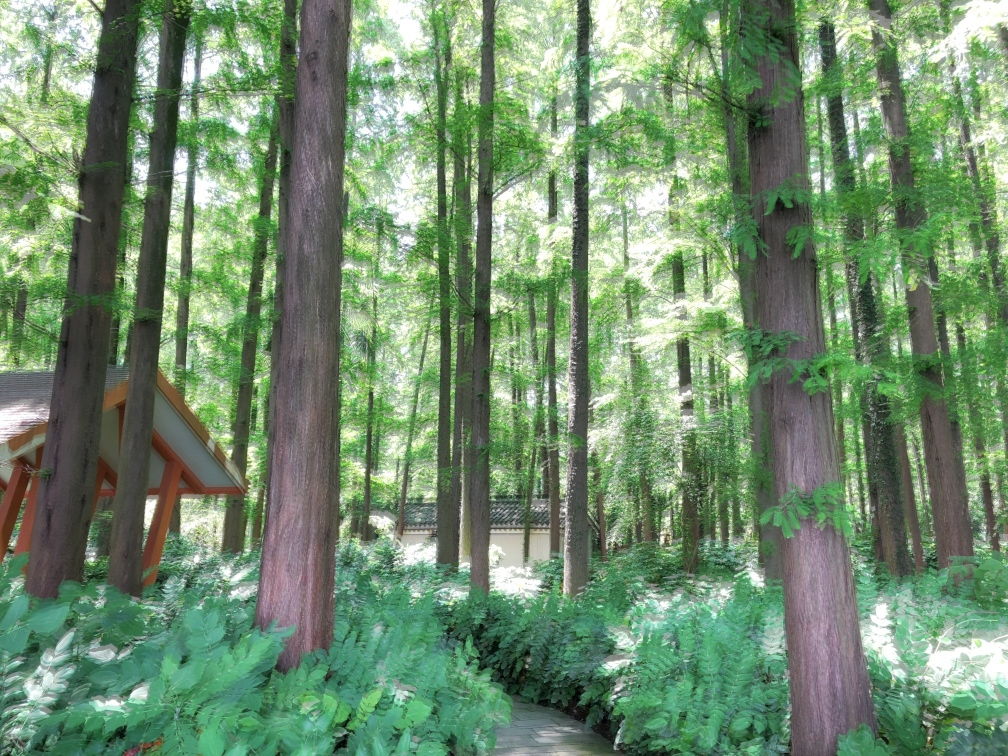}
    \includegraphics[width=.19\linewidth]{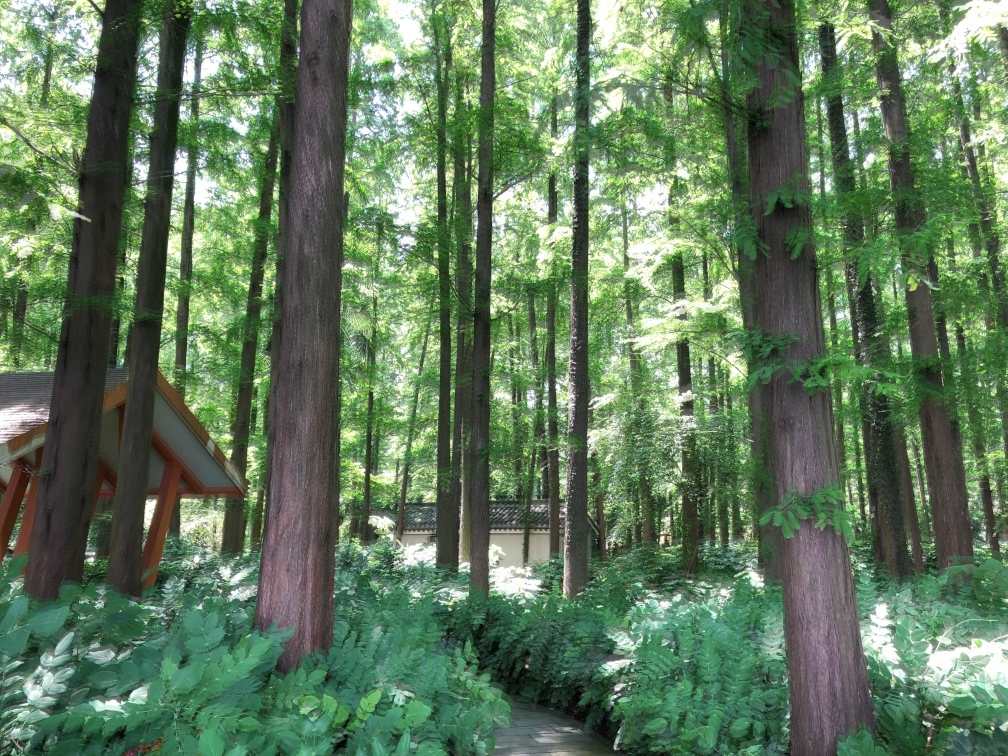}
    \includegraphics[width=.19\linewidth]{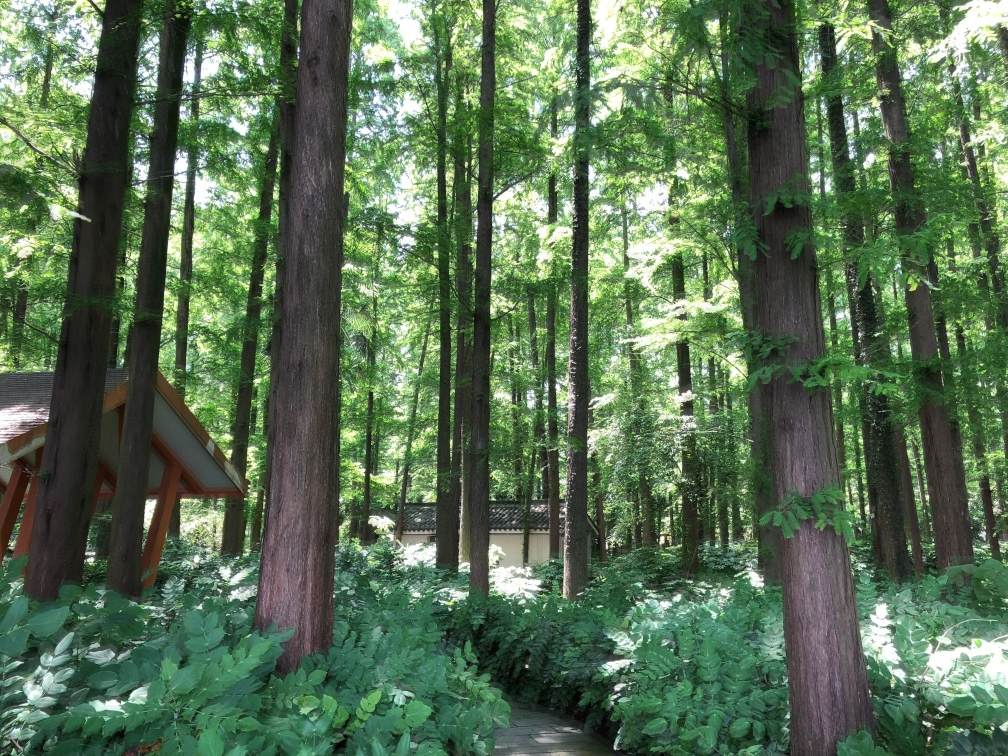}
    \includegraphics[width=.19\linewidth]{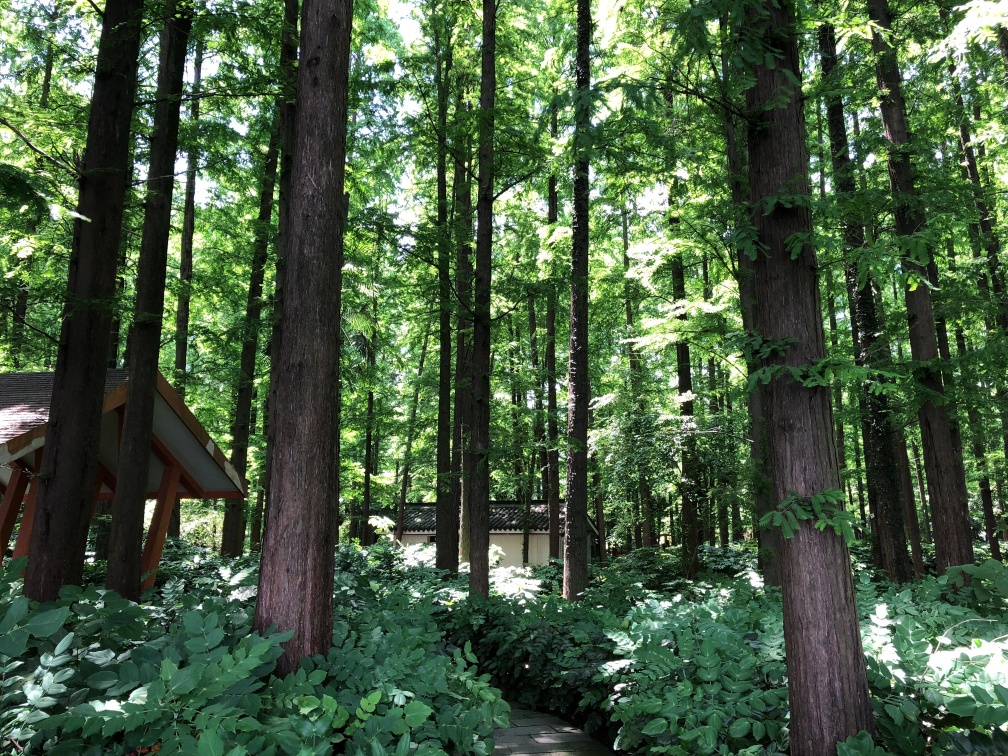}\\
    \makebox[0.19\textwidth]{\small Input}
    \makebox[0.19\textwidth]{\small FECNet~\cite{huang2022fecnet}}
    \makebox[0.19\textwidth]{\small NamedCurves~\cite{serrano2025namedcurves}}
    \makebox[0.19\textwidth]{\small Ours}
    \makebox[0.19\textwidth]{\small Ground truth}
    \makebox[\textwidth]{\small (b) SICE dataset qualitative results.}

    \caption{Qualitative results on exposure correction: (a) ME, and (b) SICE (d) datasets.}
  \label{fig:qualitative2}
\end{figure*}

Figures~\ref{fig:qualitative1} and \ref{fig:qualitative2} present qualitative comparisons across the MIT5K~\cite{bychkovsky2011mit5k}, PPR10K~\cite{liang2021ppr10k}, ME~\cite{afifi2021msec}, and SICE~\cite{cai2018sice} datasets, respectively. In Figure~\ref{fig:qualitative1} (a) and (b), we compare our framework with BGLUT~\cite{kim2025bilateral} and NamedCurves~\cite{serrano2025namedcurves} for photo retouching. The results demonstrate that our framework achieves visually appealing outcomes that most closely resemble the expert-retouched versions in terms of color fidelity. This can be observed in the blue sky in the first row, the white petals in the second row, the white dress in the third row, and the red dress in the fourth row. Figure~\ref{fig:qualitative2} (a) and (b) compare our method against FECNet~\cite{huang2022fecnet} and NamedCurves~\cite{serrano2025namedcurves} for exposure correction. Our framework consistently achieves the most accurate exposure levels, as evidenced by its ability to properly balance details in the mountains in the first row, the sky in the second row, the grass in the third row, and the light filtering through the leaves in the last row. \rev{For exposure correction, NamedCurves+ is particularly well-suited, as the backbone first normalizes the dynamic range into a canonical space, after which the color-specific tone curves independently refine each chromatic region. This two-stage approach proves more stable than single-pass methods.}

\subsection{User Interactivity}\label{subsec:interactivity}
Mimicking the style of an expert may not always align with individual user preferences. By associating tone curves with color names, NamedCurves+ provides both interpretability and interactivity. The proposed method enables users to adjust the estimated control points of the tone curves, thereby manipulating specific color regions in the final enhanced image. When a user modifies a tone curve, they directly alter all pixels with a high probability of belonging to the corresponding color region, as determined by the color naming method. Importantly, the model’s weights remain frozen during inference, ensuring that the user interactions are limited to modifying the tone curves without affecting the learned parameters.

\begin{figure*}[t]
    \centering
    \includegraphics[width=.18\linewidth]{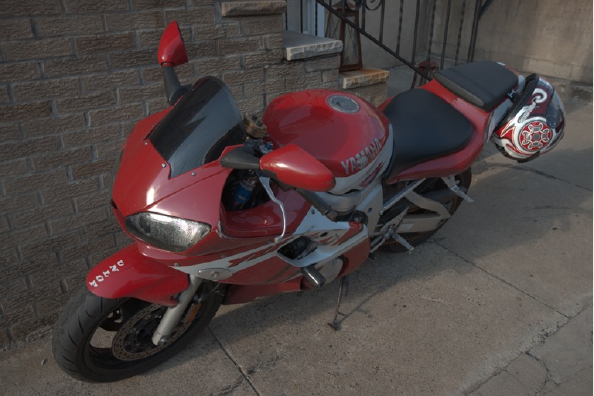}
    \includegraphics[width=.18\linewidth]{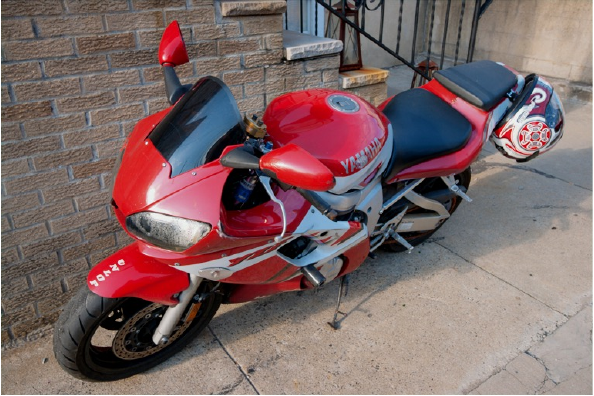}
    \includegraphics[width=.114\linewidth]{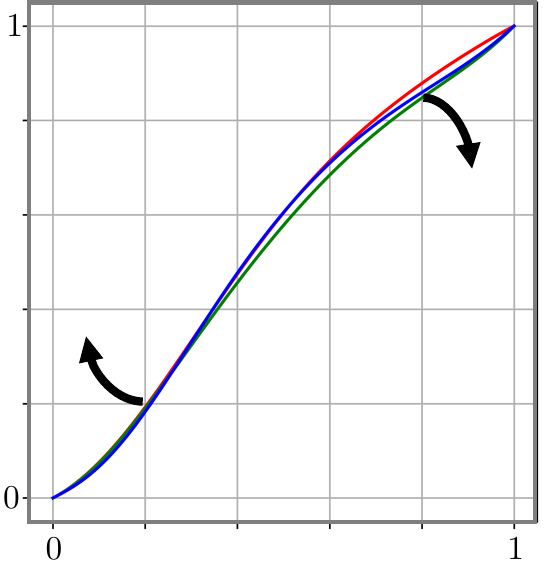}
    \includegraphics[width=.18\linewidth]{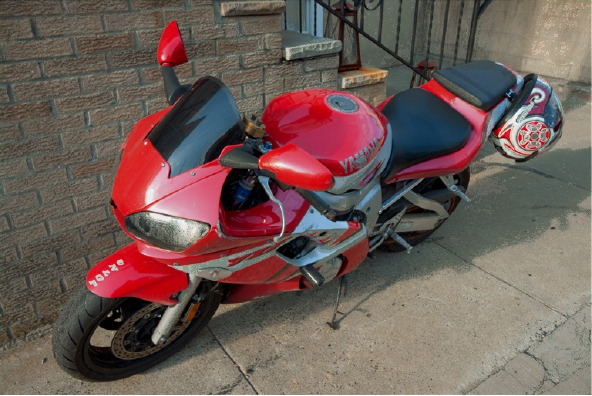}
    \includegraphics[width=.114\linewidth]{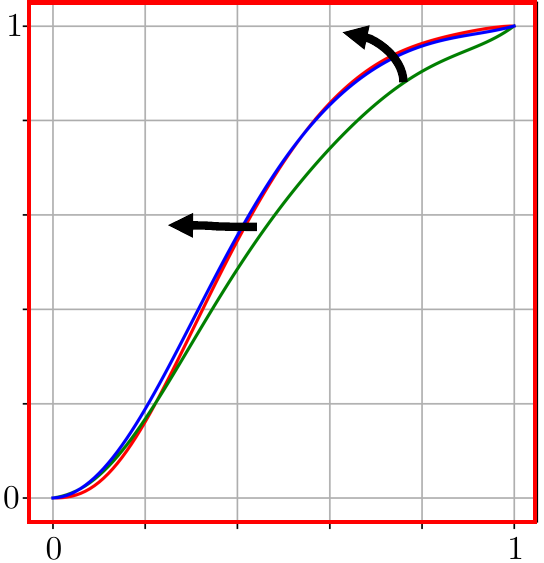}
    \includegraphics[width=.18\linewidth]{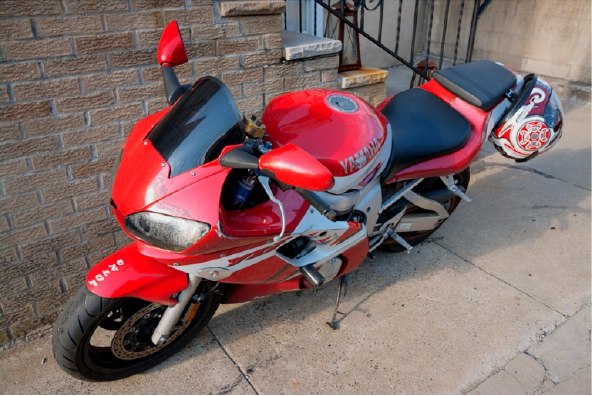}

    \makebox[.18\textwidth]{\footnotesize Input image $\hat{\text{y}}$}
    \makebox[.18\textwidth]{\footnotesize Output image $\hat{\text{y}}$}
    \makebox[.30\textwidth]{\footnotesize Modifying the Achromatic curves}
    \makebox[.30\textwidth]{\footnotesize Modifying the Red curves}

    \vspace{1mm}

    \includegraphics[width=.18\linewidth]{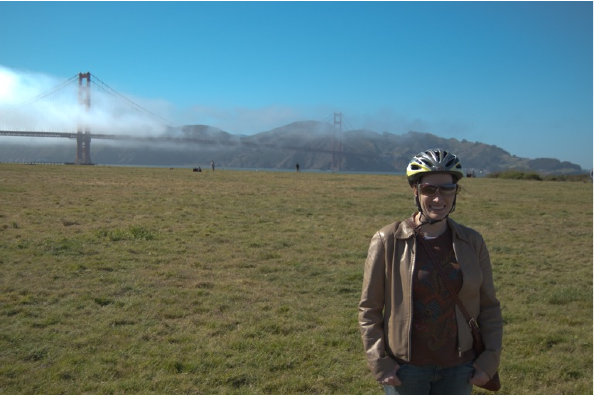}
    \includegraphics[width=.18\linewidth]{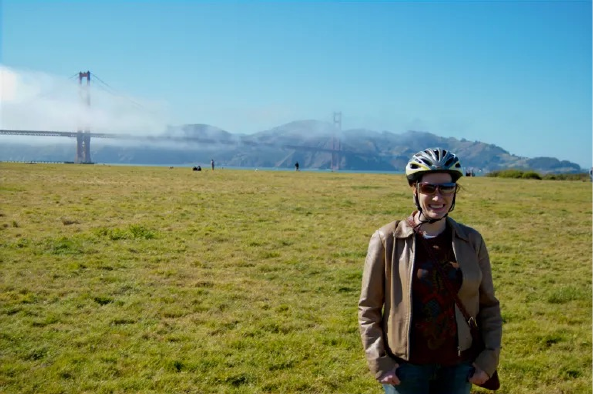}
    \includegraphics[width=.114\linewidth]{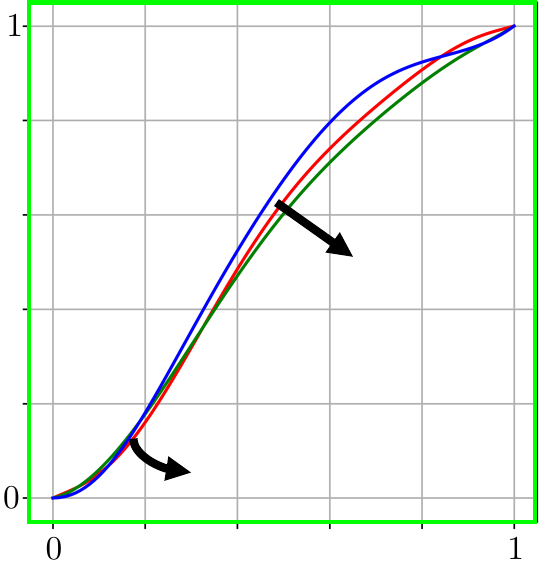}
    \includegraphics[width=.18\linewidth]{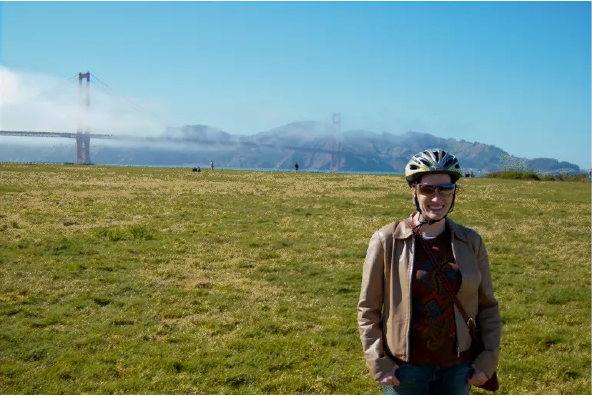}
    \includegraphics[width=.114\linewidth]{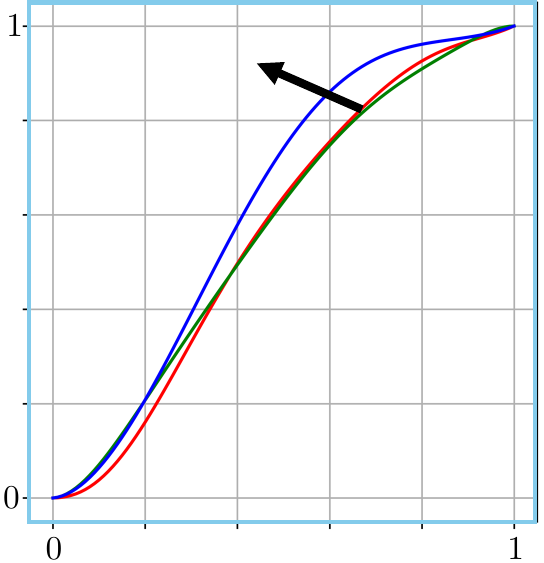}
    \includegraphics[width=.18\linewidth]{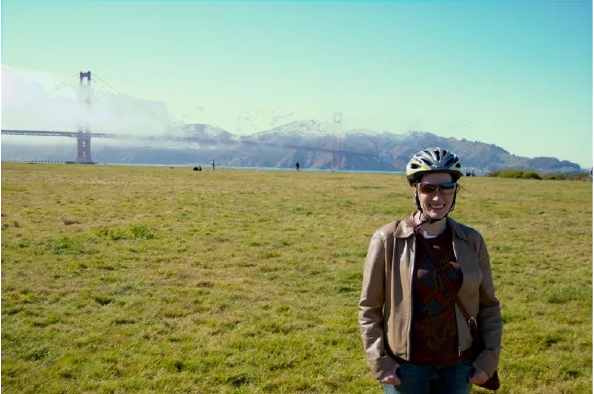}

    \makebox[.18\textwidth]{\footnotesize Input image $\hat{\text{y}}$}
    \makebox[.18\textwidth]{\footnotesize Output image $\hat{\text{y}}$}
    \makebox[.30\textwidth]{\footnotesize Modifying the Green curves}
    \makebox[.30\textwidth]{\footnotesize Modifying the Blue curves}

    \caption{Examples demonstrating the interactivity of our method. The first column displays the input image followed by the output image given by our end-to-end model. On the right, we present two distinct examples of curve modifications for each image, with black arrows indicating the direction of the adjustments.}
    \label{fig:interactive}
\end{figure*}

Figure~\ref{fig:interactive} illustrates four examples of curve modifications applied to two different images. The first two columns show the input image and the enhanced version as output by our method, respectively. We display the modified curves alongside the corresponding altered output images on the right. Black arrows indicate the direction of the curve modifications applied. In the first image, adjusting the achromatic curves reduces the shadows and highlights of the asphalt and the gray wall behind the motorbike. However, if the highlights of the red curves are increased, the red parts of the motorbike are intensified without affecting the background. In the second image, the grass or the sky is selectively intensified based on whether the green or blue curves are manipulated. We are committed to publicly releasing a demo to show our method's interactive and user-friendly nature.

\begin{figure*}[t]
    \centering
    \includegraphics[width=.19\linewidth]{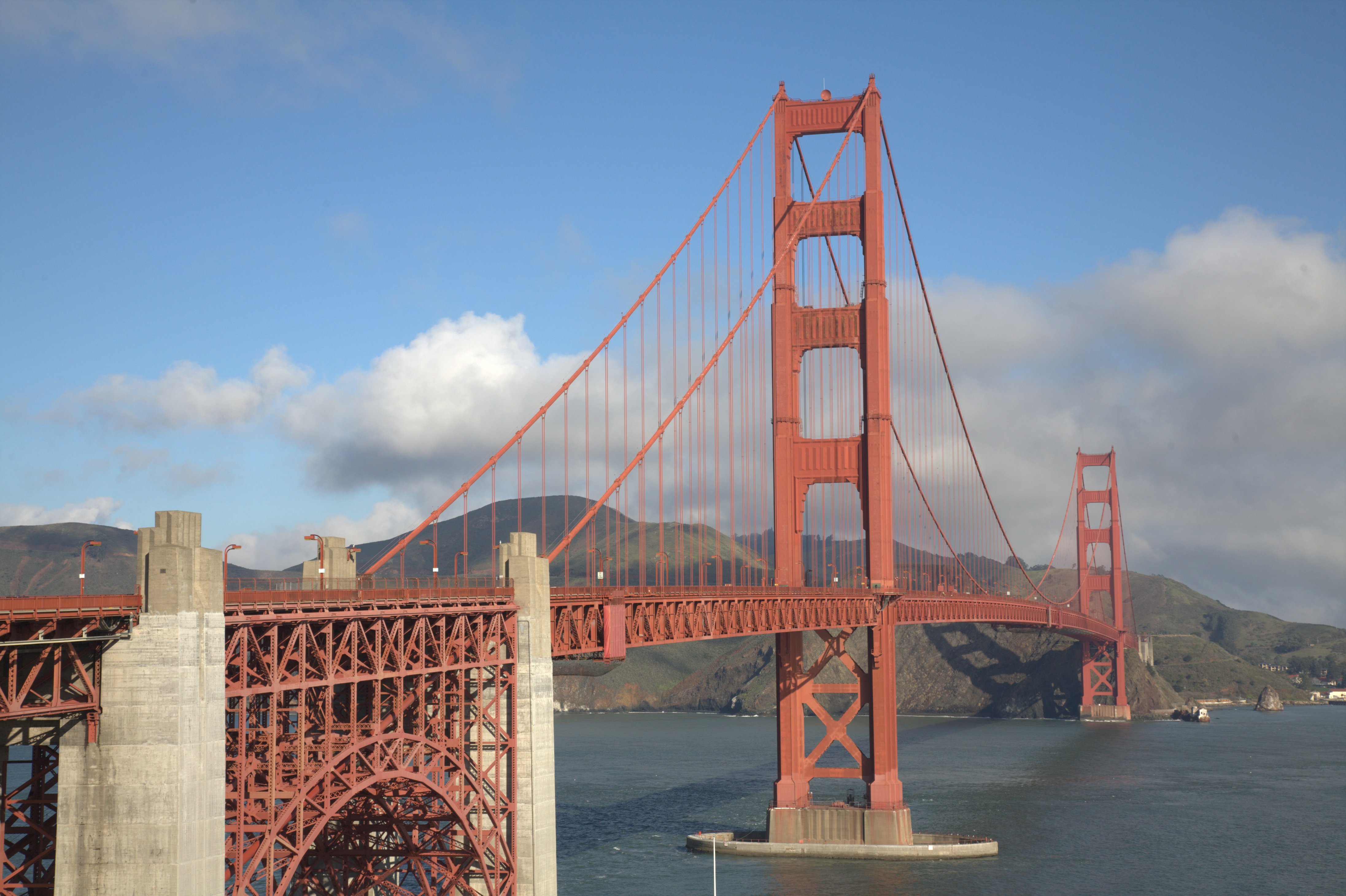}    \includegraphics[width=.19\linewidth]{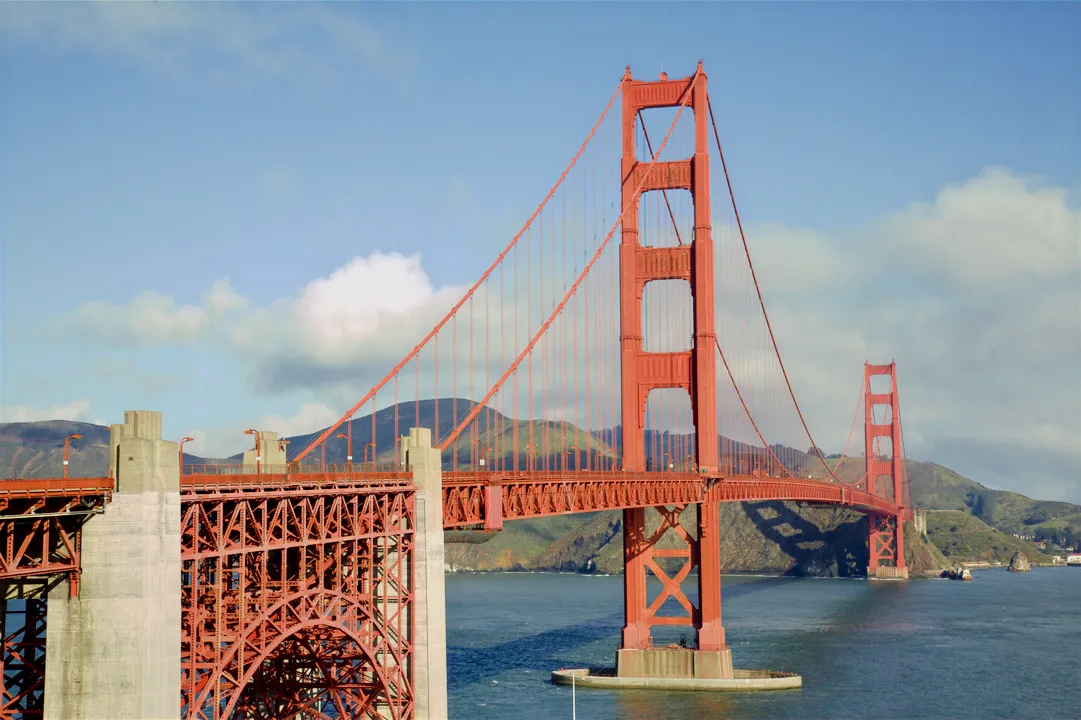}    \includegraphics[width=.19\linewidth]{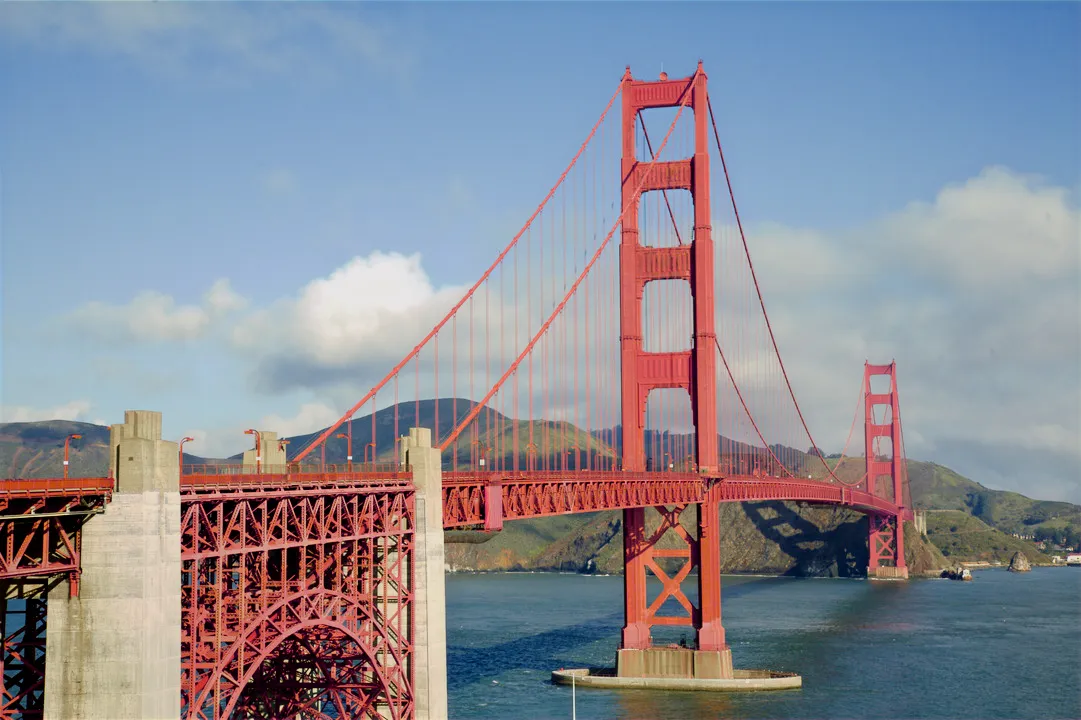}    \includegraphics[width=.19\linewidth]{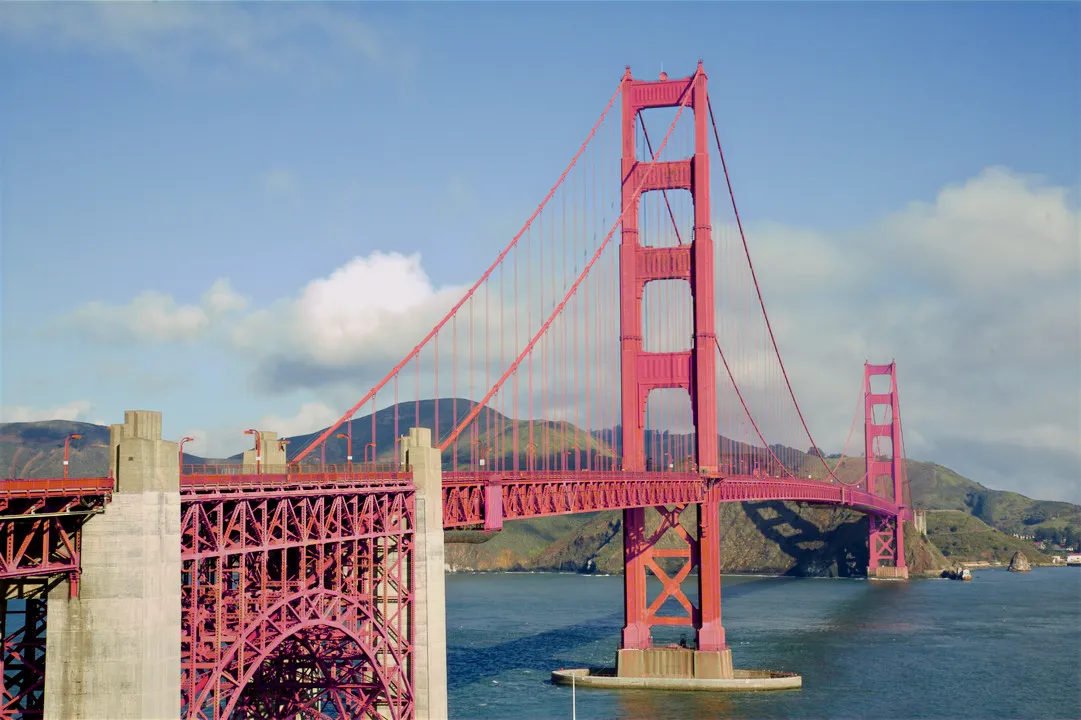}
    \includegraphics[width=.19\linewidth]{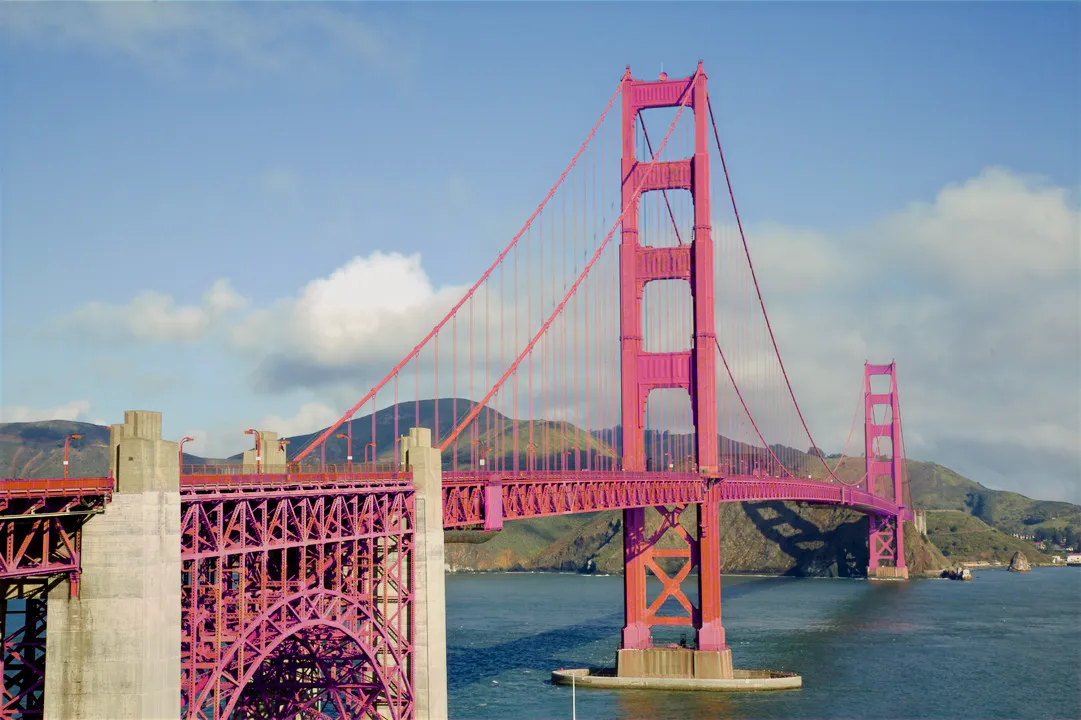}
    \caption{\rev{Progressive interactive editing of a single image. From left to right: input, NamedCurves+ automatic enhancement, and successive edits increasing all blue tone curve control points of the color name \textit{Red} by 10\% per step (when possible).}
    \vspace{-2mm}
    \label{fig:extreme_interactivity}}
\end{figure*}

\rev{Figure~\ref{fig:extreme_interactivity} demonstrates the model's ability to handle 
extreme interactive edits on the same image. The leftmost image shows the input, and the second image shows the model's automatic enhancement as a reference baseline. Then, the sequence shows progressive modifications where each step (left to right) increases all control points of the blue tone curve in the \textit{Red} channel by 10\%. With each successive edit, the red regions (most prominently the bridge) progressively shift toward a purplish hue. Notably, even under extreme and unrealistic edits, the model produces well-fused results.}

\subsection{Ablation Study on Module Contribution}
We analyze the contribution of the main components of NamedCurves+ on two representative tasks: image retouching (MIT-Adobe-5K) and exposure correction (MSEC). Table~\ref{tab:module_ablation} summarizes the effect of removing or simplifying each module.

Removing the backbone (``w/o Backbone'') causes the largest degradation, highlighting that the standardization stage is critical to map the input to a canonical space where curve-based enhancement is effective. Using only one curve and without using color naming decomposition (``w/o naming'') also decreases performance, showing that conditioning the tone curves on color terms is important to apply different adjustments to different chromatic regions. Removing the transformer (``w/o Transformer'') hurts performance as well, indicating that local, context-aware fusion is necessary beyond global edits alone. Overall, the full NamedCurves+ pipeline achieves the best results, and the improvements over NamedCurves confirm the benefit of the proposed upgrades.

\begin{table}[t!]
  \caption{\rev{Quantitative backbone analysis.}}
  \label{tab:backbone_analysis}
  \centering
  \color{myblack}
  \setlength{\tabcolsep}{3.6pt}
  \begin{tabular}{lcccccc}
    \toprule
    & \multicolumn{4}{c}{MIT-5K} & \multicolumn{2}{c}{MSEC} \\
    \cmidrule(r){2-5}     \cmidrule(l){6-7}
    \multicolumn{1}{c}{\centering Method} & PSNR & SSIM & LPIPS & $\Delta E_{00}$ & PSNR & SSIM \\
    \midrule
    Conv-UNet & 25.17 & 0.931 & 0.041 & 7.05 & 22.21 & 0.864\\
    TED & 25.22 & 0.934 & 0.039 & 6.87 & 22.45 & 0.868\\
    Restormer & 25.47 & 0.938 & 0.036 & 6.33 & 22.94 & 0.873\\
    LPIENet & 25.51 & 0.938 & 0.035 & 6.29 & 22.98 & 0.874\\
    Ours & 25.75 & 0.940 & 0.035 & 6.01 & 23.15 & 0.878\\
  \bottomrule
  \end{tabular}
  \vspace{-3mm}
\end{table}

\begin{table}[t!]
  \caption{Quantitative comparisons on the MIT-Adobe-5K dataset~\cite{bychkovsky2011mit5k}.}
  \centering
  \setlength{\tabcolsep}{3.6pt}
  \begin{tabular}{lcccccc}
    \toprule
    \multicolumn{1}{c}{\centering Method} & Channels & PSNR & SSIM & LPIPS & $\Delta E_{00}$ & $\Delta E_{ab}$\\
    \midrule
    w/o color naming & 1 & 24.91 & 0.927 & 0.056 & 6.67 & 7.74 \\
    w/o color naming & 6 & 25.04 & 0.928 & 0.058 & 6.70 & 7.76 \\
    w/o color naming & 11 & 24.83 & 0.927 & 0.055 & 6.67 & 7.69 \\
    \midrule
    RGB cube vertices & 8 & 24.96 & 0.928 & 0.059 & 6.74 & 7.81 \\
    Random GMM & 11 & 24.09 & 0.931 & 0.054 & 6.59 & 7.63 \\
    Random GMM & 6 & 24.12 & 0.931 & 0.052 & 6.56 & 7.62 \\
    \midrule
    Benavente et al.~\cite{benavente2008parametric} & 11 & 25.32 & 0.930 & 0.049 & 6.38 & 7.48 \\
    Benavente et al.~\cite{benavente2008parametric} & 6 & 25.57 & 0.935 & 0.039 & 6.18 & 7.39 \\
    \midrule
    Van de Weijer~\cite{van2009learning} & 11 & 25.49 & 0.932 & 0.041 & 6.28 & 7.46 \\
    Van de Weijer~\cite{van2009learning} & 6 & \textbf{25.75} & \textbf{0.940} & \textbf{0.035} & \textbf{6.01} & \textbf{7.27} \\
  \bottomrule
  \end{tabular}
  \label{tab:color_naming}
\end{table}

\begin{figure}[h]
  \centering
  \includegraphics[width=\linewidth]{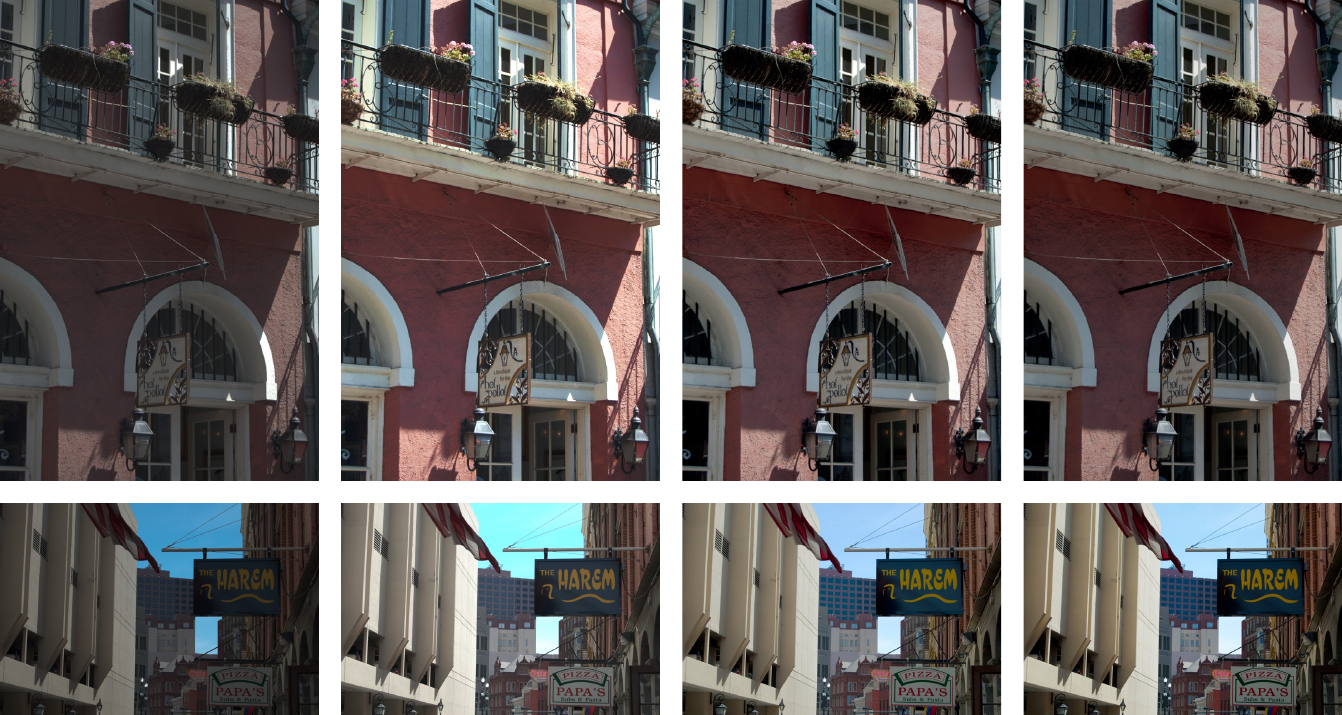}
  \makebox[.24\linewidth]{\footnotesize Input image}
  \makebox[.24\linewidth]{\footnotesize Standardized image}
  \makebox[.24\linewidth]{\footnotesize NamedCurves+}
  \makebox[.24\linewidth]{\footnotesize Ground Truth}
  \vspace{-4mm}
  \caption{\rev{Examples of NamedCurves+ and the intermediate standardized image. The standardized image compresses the dynamic range and corrects exposure before color-specific curves are applied.}}
  \vspace{-6mm}
  \label{fig:standardized_comparison}
\end{figure}

\subsection{Backbone Analysis}

The backbone produces the standardized image $\hat{y}_b$, which serves as the starting point for the subsequent curve-based global editing and transformer-based local fusion. To better understand the impact of this standardization stage, we replace the backbone with several representative architectures and evaluate the overall pipeline while keeping the rest of the framework unchanged. Table~\ref{tab:backbone_analysis} reports the results on MIT-Adobe-5K and MSEC. We observe that stronger restoration-oriented backbones consistently lead to better final enhancement quality, especially on the more challenging exposure correction setting (MSEC), where recovering details in shadows/highlights is essential. While a Conv-UNet already provides a reasonable baseline, transformer-based or hybrid designs (e.g., Restormer and LPIENet) improve perceptual similarity and reduce color error. Our backbone achieves the best overall performance across datasets and metrics, indicating that accurate standardization is a key enabler: it reduces the burden on the curve estimation module (which can then focus on interpretable, color-conditioned global edits) and provides cleaner inputs for the transformer fusion module to model local, context-dependent corrections.

In Figure~\ref{fig:standardized_comparison}, we show two examples of NamedCurves+, and the intermediate standardized image $\hat{y_b}$. By imposing an additional constraint on the loss term with $\hat{y_b}$, we impose that the standardized image is a first enhanced version. As can be seen in Figure~\ref{fig:standardized_comparison}, the standardized image corrects the exposure of the input image. Then the color naming curves and fusion module refine it to obtain closer results to the ground truth.

\begin{table}[htbp]
    \centering
    \caption{\rev{Computational cost comparison.}}
    \color{myblack}
    \begin{tabular}{lcccc}
    \toprule
     Fusion Module & Inf. time & Num. Params & VRAM & FLOPs \\
     \midrule
     NamedCurves & 26ms & 993K & 2.44GB & 814G \\
     NamedCurves+ & 19ms & 1015K & 3.40GB & 766G \\
     \bottomrule
    \end{tabular}
    \label{tab:computational}
\end{table}

\begin{table}[htbp]
\centering
\begin{minipage}{0.48\linewidth}
    \centering
    \setlength{\tabcolsep}{5pt}
    \caption{Study on the number of control points. MIT5K}
    \label{tab:points_ablation}
    \begin{tabular}{cccc}
        \toprule
        N & PSNR & SSIM & $\Delta E_{00}$\\
        \midrule
        7  & 25.70 & 0.938 & 6.04 \\
        11 & \textbf{25.75} & \textbf{0.940} & \textbf{6.01} \\
        16 & 25.59 & 0.935 & 6.15 \\
        \bottomrule
    \end{tabular}
\end{minipage}
\begin{minipage}{0.48\linewidth}
    \centering
    \setlength{\tabcolsep}{5pt}
    \caption{Study on $\alpha$ of the loss function. MIT5K}
    \label{tab:alpha_ablation}
    \begin{tabular}{cccc}
        \toprule
        $\alpha$ & PSNR & SSIM & $\Delta E_{00}$\\
        \midrule
        0  & 25.41 & 0.930 & 6.29 \\
        0.5 & \textbf{25.75} & \textbf{0.940} & \textbf{6.01} \\
        1 & 25.58 & 0.937 & 6.21 \\
        \bottomrule
    \end{tabular}
\end{minipage}
\end{table}

\subsection{Color Naming Ablation Study}
In this section, we perform an ablation study to assess the importance of color naming and its impact on the performance of our method. Table~\ref{tab:color_naming} presents the results of all ablation experiments, where we report PSNR, SSIM, LPIPS, $\Delta E_{00}$, and $\Delta E_{ab}$. We compare the color naming methods proposed by Van de Weijer et al.~\cite{van2009learning} and Benavente et al.~\cite{benavente2008parametric}, both employing 11 color names, as well as our proposed grouping strategy to reduce them to six color categories. Additionally, we experimented with a simple baseline method in which a pixel probability assignment was based on the distance to the vertices of the RGB cube, resulting in eight channels. To further analyze the role of color grouping, we also generate six and eleven random classes using Gaussian Mixture Models (GMM) over the RGB color space. Finally, we evaluate our method without conditioning the curve estimation process with the color naming decomposition. The results show that, among the different partitioning methods evaluated (i.e., random GMM, RGB cube vertices), color naming is the most adequate to model enhancement operations since color names correlate with human perception, supporting our initial hypothesis. Among color naming methods, Van de Weijer et al.~\cite{van2009learning} outperforms Benavente et al.~\cite{benavente2008parametric}. Moreover, reducing the number of color terms to six yields better performance than using all 11 color terms individually.

The core module of NamedCurves+ is the color naming decomposition. We provide evidence of its orthogonality and applicability to other backbones and architectures through two existing experiments. First, our backbone ablation (Table~\ref{tab:backbone_analysis}) shows that the color naming head consistently improves performance across different architectural backbones, demonstrating that the semantic decomposition yields complementary gains independent of the underlying estimation mechanism. Second, our color naming ablation (Table~\ref{tab:color_naming}) shows that color-based decomposition substantially outperforms alternative partitioning strategies, achieving 25.75 dB PSNR compared to 24.96 dB for RGB cube vertices and 24.12 dB for random GMM. This confirms that the gains originate from the semantic guidance itself, rather than from the fusion mechanism alone. Taken together, these results strongly support the orthogonality of our contribution.

\subsection{Fusion Mechanism Computational Cost}
Table~\ref{tab:computational} presents a detailed computational analysis comparing our new fusion module with the previous approach. We report: (i) average inference time over 100 images from the MIT5K dataset on an AMD EPYC 7642 CPU and NVIDIA A40 GPU, (ii) total number of parameters for both complete models, (iii) peak VRAM consumption, and (iv) FLOPs for processing a 1024$\times$1024 image. The new Transformer fusion mechanism adds minimal overhead in terms of parameters (+22K, or 2.2\% increase) while requiring moderately higher VRAM (+0.96GB) due to full-resolution processing. Importantly, despite these modest increases, the approach achieves both faster inference (27\% reduction from 26ms to 19ms) and reduced computational cost (6\% fewer FLOPs, from 814G to 766G), while simultaneously improving results both quantitatively and qualitatively. The efficiency gains stem from replacing six independent attention modules with a single unified Transformer block that leverages GPU-optimized transposed attention operations.

\subsection{Hyperparameter Analysis}

We conducted additional analysis to evaluate the impact of the number of control points, $N$, and the parameter $\alpha$ in the loss function. Specifically, we tested our method with 7, 11, and 16 control points for each Bézier curve. Table~\ref{tab:points_ablation} reports the PSNR, SSIM, and $\Delta E_{00}$ for each configuration. Our experiments show that 11 control points, spaced evenly every 0.1 along the input axis, yield the best performance. Additionally, we examined the influence of the $\alpha$ parameter, which balances the contribution of the backbone and the final image component in the loss function. We tested three settings: removing the backbone component ($\alpha{=}0$), assigning it half the weight of the final image component ($\alpha{=}0.5$), and giving it equal weight ($\alpha{=}1$). As shown in Table~\ref{tab:alpha_ablation}, we found that $\alpha{=}0.5$ provides the best results for our method.

\subsection{User Study}

\begin{figure}
    \centering
    \includegraphics[width=\linewidth]{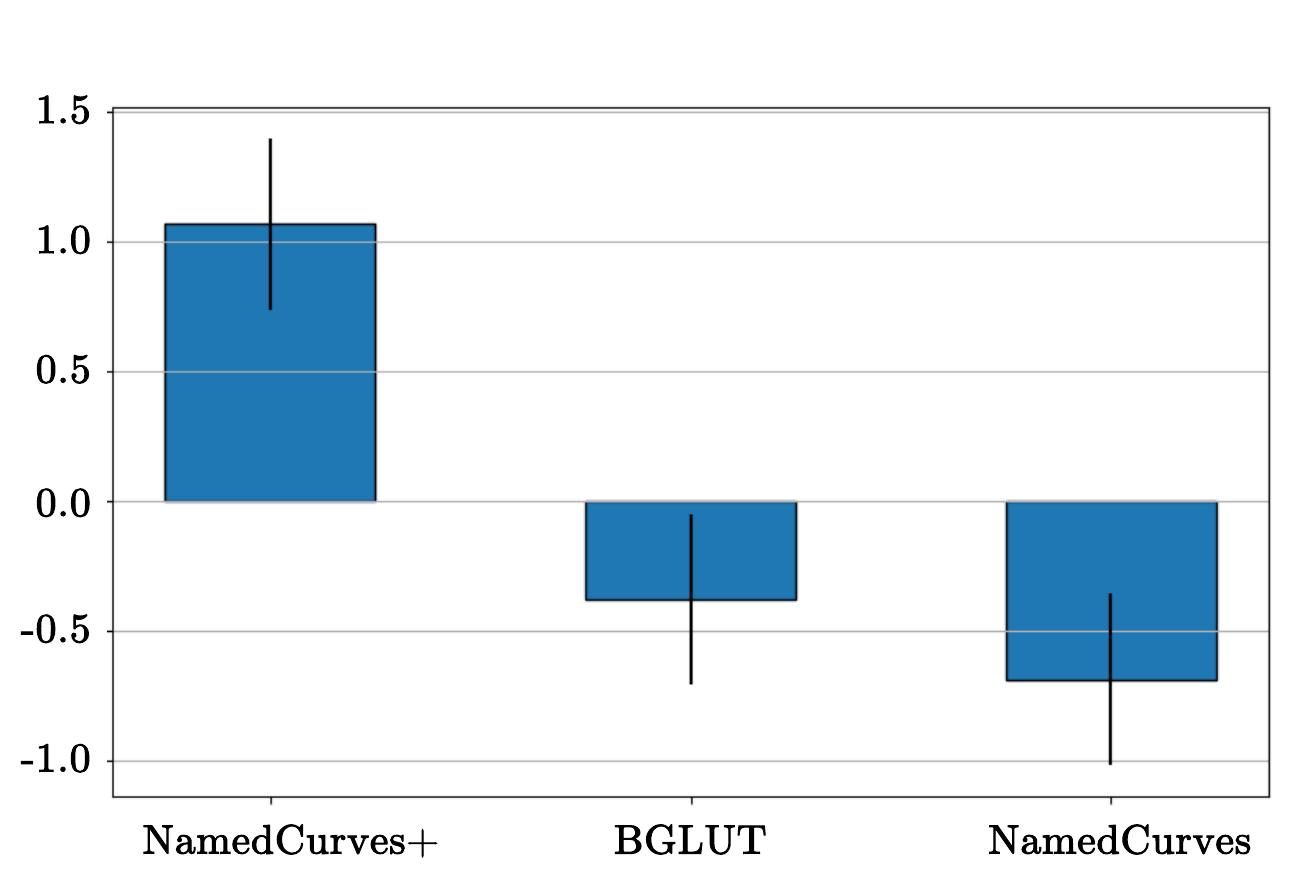}
    \vspace{-4mm}
    \caption{\rev{Two-alternative forced choice (2AFC) on 25 images from MIT5K and PPR10K datasets. 15 observers took part.}}
    \label{fig:user_study}
\end{figure}

We compared NamedCurves+ against NamedCurves and BGLUT following a two-alternative forced choice (2AFC), performed in a completely black room with a monitor set to sRGB. We selected the same 25 images from both MIT5K and PPR10K datasets as in~\cite{serrano2025namedcurves} (that were selected randomly). 15 observers (11M/4F; ages 20-34) participated in the study, and all were screened for colour vision deficiency using the Ishihara test. Results analyzed using Thurstone Case V (larger means better) were: NamedCurves+: 1.0672; BGLUT: -0.3793; NamedCurves: -0.6878. Our method is statistically significantly better than the other two —95\% confidence interval is 0.33. Figure~\ref{fig:user_study} shows the results.

\section{Limitations}\label{sec:limitations}

Despite the strong quantitative results and the interpretability afforded by color-specific tone curves, our approach has some limitations.
First, the global editing space is constrained by the expressiveness of tone curves applied per color term. This is well suited to global color and contrast changes, but it cannot directly represent highly localized or semantic operations (e.g., object-specific relighting, instance-level recoloring, or edits that depend on high-level scene understanding). The transformer fusion alleviates this limitation by incorporating local context, but it still operates on a fixed set of globally manipulated candidates, and thus remains bounded by them.
Second, the approach depends on the quality of the color naming decomposition. When color term probabilities are uncertain (e.g., at chromatic boundaries, saturated highlights, or under unusual illuminants), edits may partially leak across regions. Finally, a limitation shared by all retouching methods that train a model to mimic the style of an \textit{expert} is the difficulty of generalizing to the preferences of all users. We partially address this by allowing users to interactively modify the color-specific tone curves; nevertheless, the automatic output will inherently reflect the aesthetic style of the training data.

\section{Conclusion}\label{sec:conclusion}
In this paper, we introduce NamedCurves+, a novel approach to efficiently enhance images in three scenarios: photo retouching, tone mapping, and exposure correction. NamedCurves+ incorporates significant advancements, including a transformer block for enhanced image fusion, support for additional tasks with more efficient computation, and analysis of interactivity. This interactivity enables users to fine-tune image colors independently by adjusting the tone curves, allowing them to customize the results to match their preferences better.

\bibliographystyle{IEEEtran}
\bibliography{sn-bibliography}

\section{Acknowledgements}
This work was supported by Grant PID2021-128178OB-I00, , PID2024-162555OB-I00 funded by MCIN/AEI/10.13039/ 501100011033 and by ERDF "A way of making Europe", the grant Catedra ENIA UAB-Cruïlla
(TSI-100929-2023-2) from the Ministry of Economic
Affairs and Digital Transformation of Spain, and by the Generalitat de Catalunya CERCA Program. DSL also acknowledges the FPI grant from Spanish Ministry of Science and Innovation (PRE2022-101525). LH was also supported by the Ramon y Cajal grant RYC2019-027020-I. The CFREF (VISTA) program, an NSERC Discovery Grant, and the Canada Research Chair program supported MSB.

\begin{IEEEbiography}
[{\includegraphics[width=1in,height=1.25in,clip,keepaspectratio]{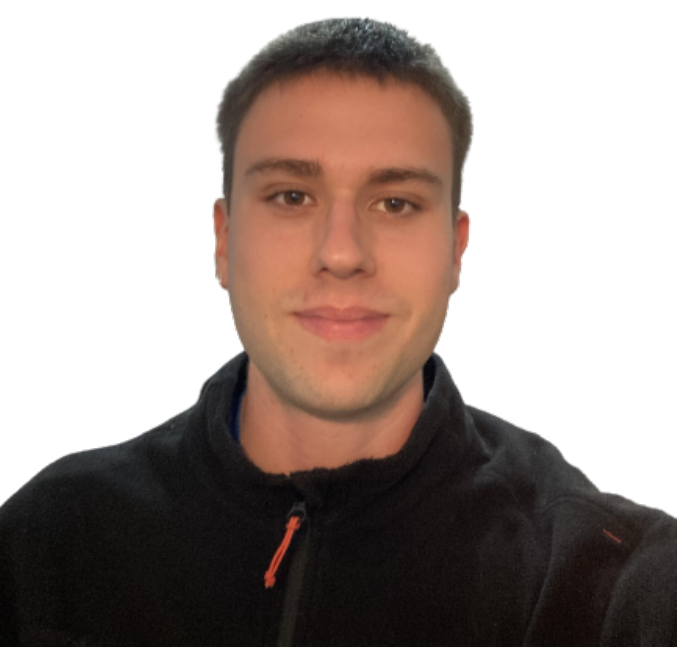}}]{David Serrano-Lozano} is a PhD student at the Computer Vision Center and Universitat Autònoma de Barcelona. He received his B.Sc. degree from Universitat Politècnica de Catalunya and his M.Sc. degree from Universitat Autònoma de Barcelona. His research interests include low-level computer vision, generative models, and computational imaging. 
\end{IEEEbiography}

\begin{IEEEbiography}
[{\includegraphics[width=1in,height=1.25in,clip,keepaspectratio]{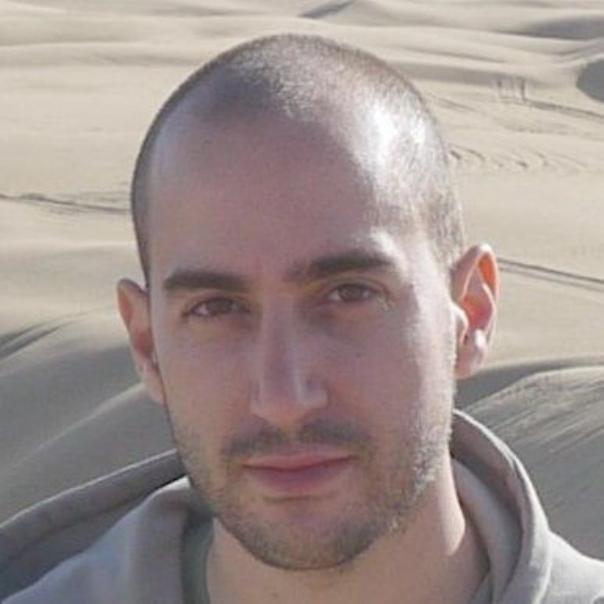}}]{Luis Herranz} is a researcher at the Universidad Autónoma de Madrid. He received his Ph.D. degree from the Universidad Autónoma de Madrid in 2010. He has worked at the Computer Vision Center (Barcelona), the Institute of Computing Technology (Chinese Academy of Sciences), and Mitsubishi Electric Research and Development. His research interests include deep learning applied to various topics in computer vision and multimedia.
\end{IEEEbiography}

\begin{IEEEbiography}
[{\includegraphics[width=1in,height=1.25in,clip,keepaspectratio]{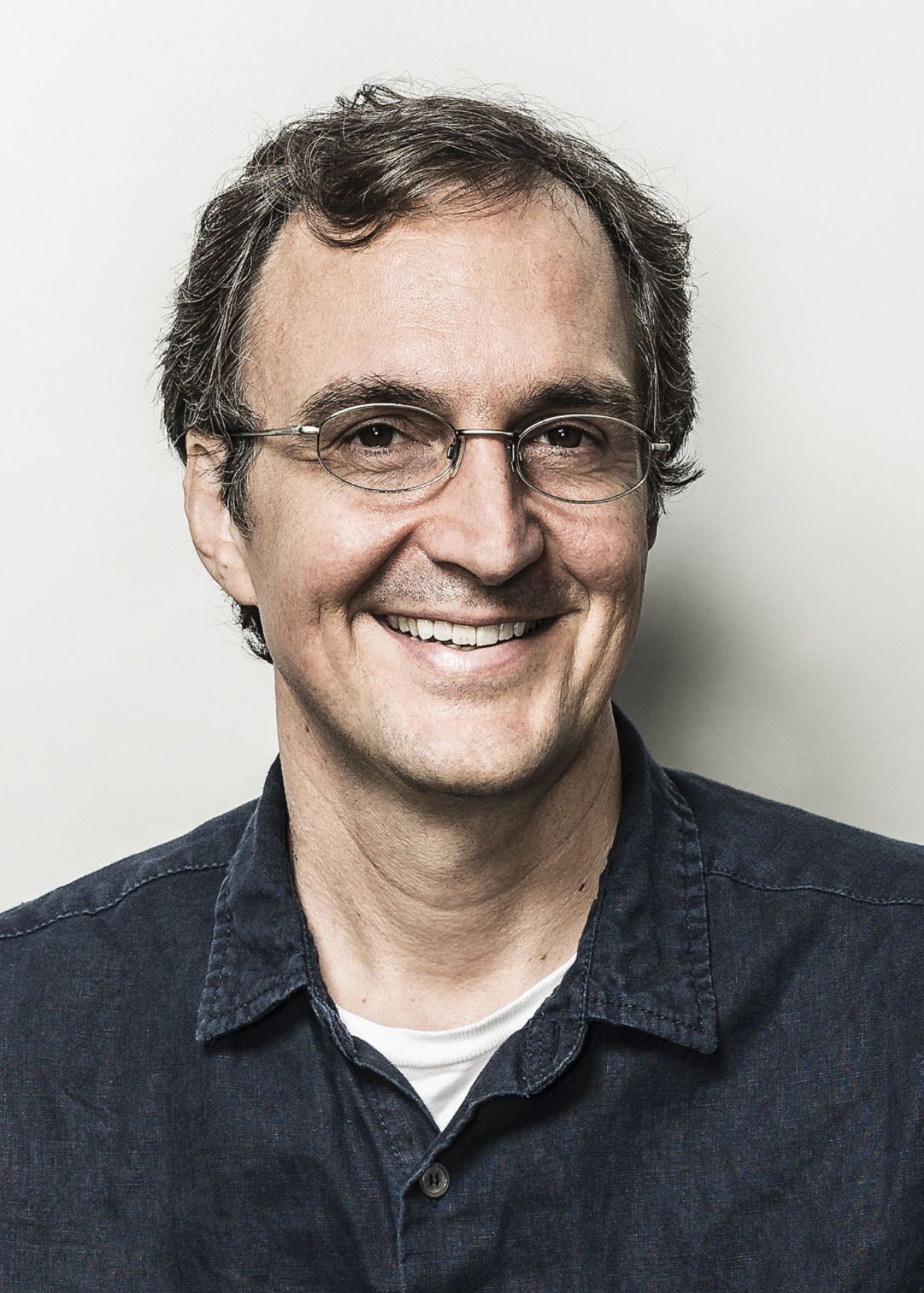}}]{Michael S. Brown} is a Professor and a Canada Research Chair of Computer Vision with York University, Toronto. He also holds a part-time position as a Senior Research Director at the Samsung AI Center, Toronto. His research interests include computer vision, image processing, and computer graphics. He has served as the Program Chair for WACV (2011, 2017, 2019) and 3DV (2015), as well as General Chair for ACCV (2014) and CVPR (2018, 2021, 2023). He is a Fellow of the IEEE.
\end{IEEEbiography}

\begin{IEEEbiography}
[{\includegraphics[width=1in,height=1.25in,clip,keepaspectratio]{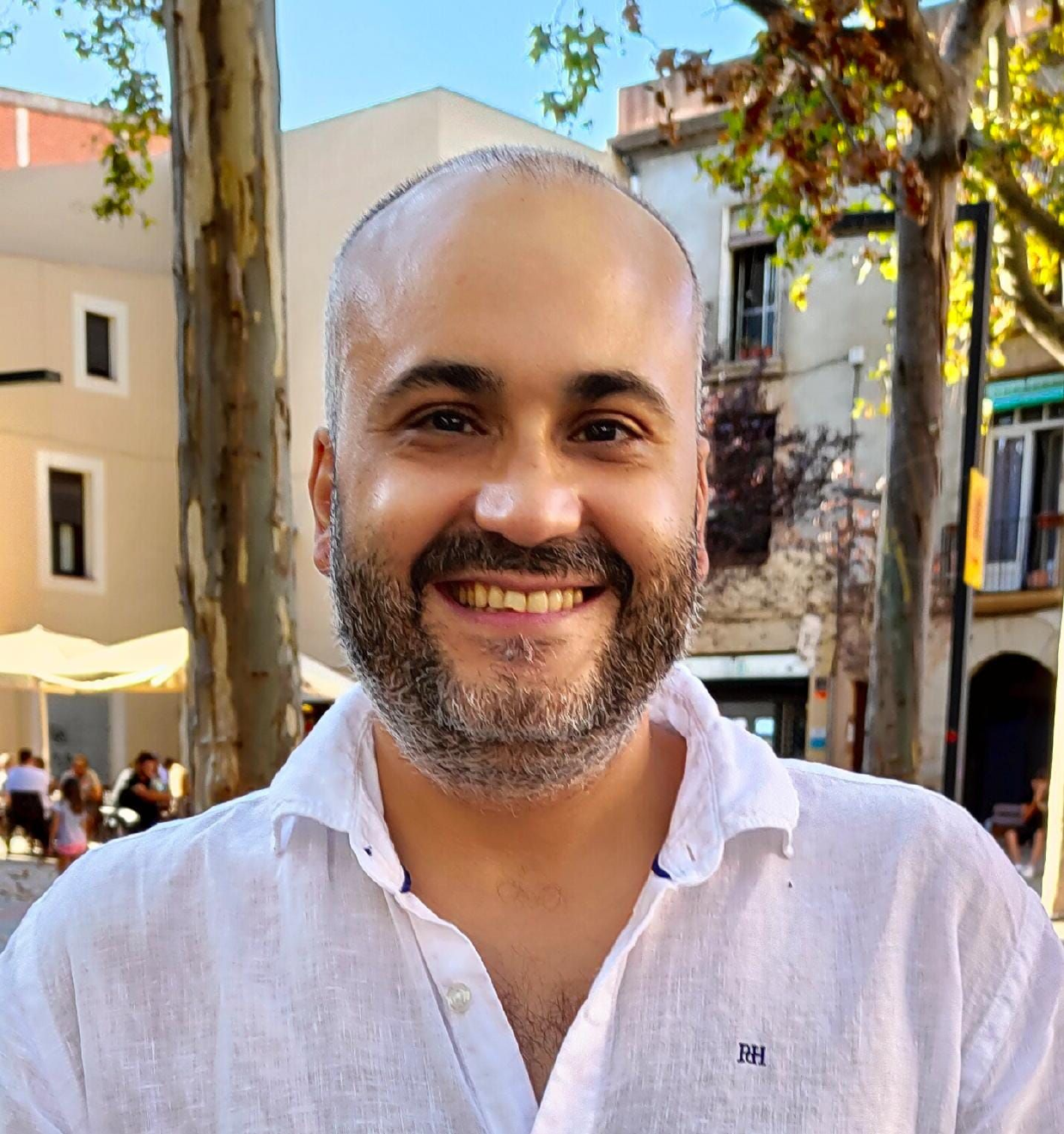}}]{Javier Vazquez-Corral} is an Associate Professor at the Universitat Autònoma de Barcelona and a researcher at the Computer Vision Center, Barcelona. He received his Ph.D. degree from the Universitat Autònoma de Barcelona. He has worked at the University of East Anglia, the École Polytechnique Fédérale de Lausanne, and the Universitat Pompeu Fabra. His research interests include the use of color in computer vision problems and bridging the gap between color perception in the human brain and its application in computer vision.
\end{IEEEbiography}

\vfill

\end{document}